\documentclass[10pt,twocolumn,letterpaper]{article}
 \usepackage[pagenumbers]{cvpr} 

\usepackage{graphicx}
\graphicspath{{../2-IEEE/},{../3-CVPR/}} 
\usepackage{amsmath}
\usepackage{amssymb}
\usepackage{booktabs}
\usepackage{xcolor}         
\usepackage{enumerate}
\usepackage[normalem]{ulem}

\usepackage{multirow}
\renewcommand{\arraystretch}{1.0} 
\usepackage{etoolbox}
\usepackage[pagebackref,breaklinks,colorlinks]{hyperref}

\usepackage[capitalize]{cleveref}
\crefname{section}{Sec.}{Secs.}
\Crefname{section}{Section}{Sections}
\Crefname{table}{Table}{Tables}
\crefname{table}{Tab.}{Tabs.}

\def\cvprPaperID{211} 
\def\confName{CVPR}
\def\confYear{2022}

\begin{document}
\title{IDEA-Net: Dynamic 3D Point Cloud Interpolation \\via Deep Embedding Alignment}





\author{ 
Yiming Zeng$^{1}$\quad Yue Qian$^1$\quad Qijian Zhang$^1$\quad Junhui Hou$^{1}$\quad Yixuan YUAN$^1$ \quad Ying He$^2$\\
$^1$
City University of Hong Kong~~ 
$^2$
Nanyang Technological University\\
{\tt 
ym.zeng
@my.cityu.edu.hk, 
jh.hou
@cityu.edu.hk  
}
}

\maketitle
\footnotetext[1]{This work was supported by the HK RGC Grant CityU 11202320 and 11218121. Corresponding author: J. Hou.}

\begin{abstract}
This paper investigates the problem of temporally interpolating dynamic 3D point clouds with large non-rigid deformation. 
We formulate the problem as estimation of point-wise trajectories (i.e., smooth curves) and further reason that temporal irregularity and 
under-sampling 
are two major challenges. 
To tackle the challenges, we propose IDEA-Net, an end-to-end deep learning framework, which disentangles the problem under the assistance  of the explicitly learned temporal consistency.
Specifically, we propose a temporal consistency learning module to align two consecutive point cloud frames point-wisely, based on which we can employ linear interpolation to obtain coarse trajectories/in-between frames. To compensate the high-order nonlinear components of trajectories,
we apply aligned feature embeddings that encode local geometry properties to regress point-wise increments, which are combined with the coarse estimations. We demonstrate the effectiveness of our method on various point cloud sequences and observe 
large improvement over state-of-the-art methods both quantitatively and visually. 
Our framework can bring benefits to 3D motion data acquisition. The source code is publicly available at \href{https://github.com/ZENGYIMING-EAMON/IDEA-Net.git}{https://github.com/ZENGYIMING-EAMON/IDEA-Net.git}.
\end{abstract}
\vspace{-0.4cm}
\section{Introduction}\label{sec:introduction}\vspace{-0.5em}

\begin{figure}[t]
\setlength{\abovecaptionskip}{0.2cm}
\setlength{\belowcaptionskip}{0.2cm}
	\centering
    \includegraphics[width=0.7\linewidth]{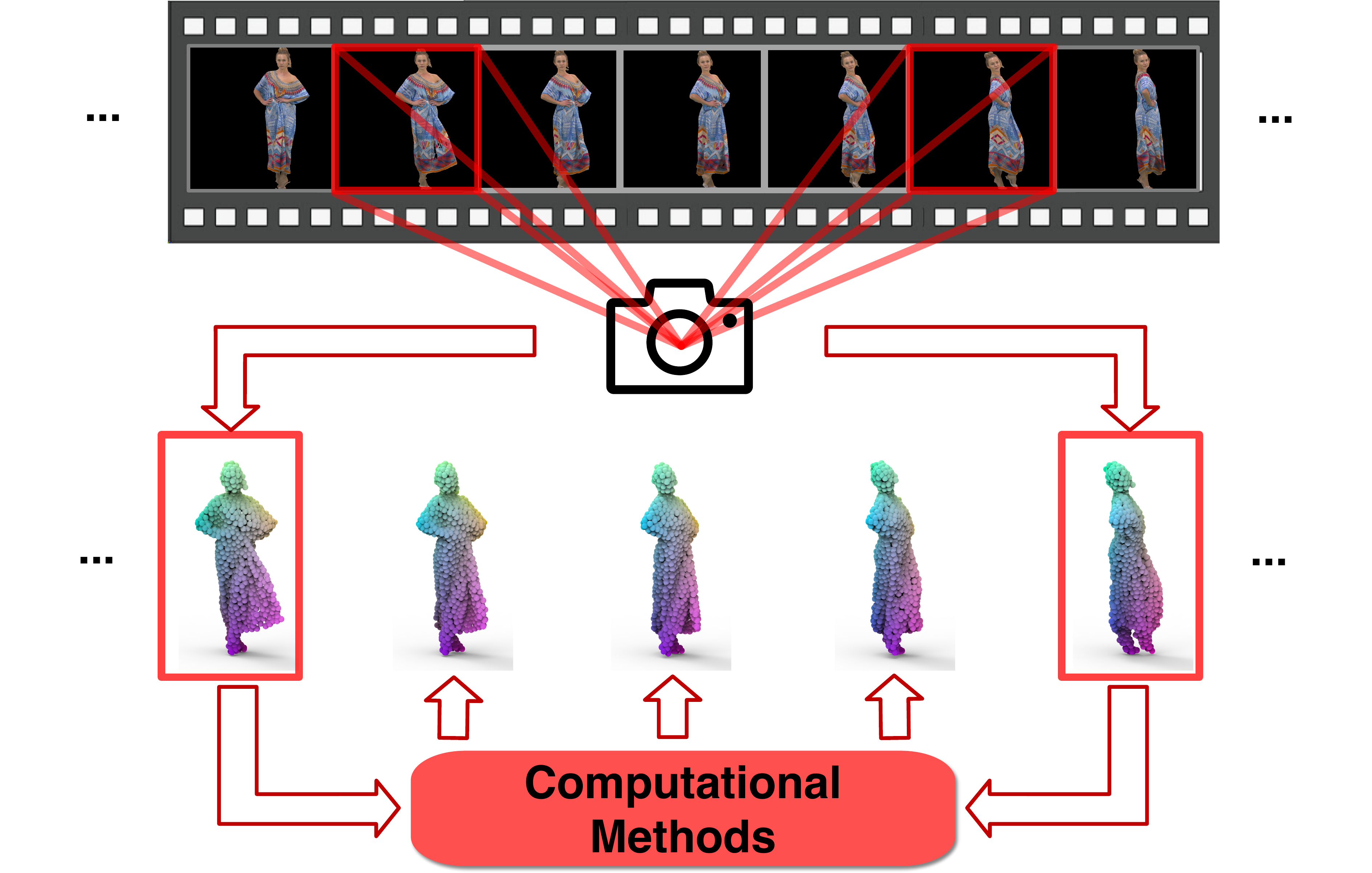}
	\caption{Illustration of the problem considered in this paper.
One can adopt a low-cost 3D sensing device to sample the motion at a low frequency, leading to an LTR point cloud sequence, then apply the computational method to interpolate/estimate in-between point cloud frames to obtain an HTR one 
for finely representing the 3D motion of objects (or 3D shapes/objects deforming over time). We are interested in point cloud sequences with \textit{massive non-rigid deformation}. Moreover, in real application scenario, the point cloud frames of a sequence are independently captured in the sensor space, thus lacking \textit{point-wise temporal consistency}. 
}
	\label{fig:scene}
\end{figure}

Dynamic 3D point clouds, which are sequences of 3D point cloud frames sampled in the temporal domain for capturing the changes in geometric details or motion of scenes/objects,  have been widely used in many application scenarios, such as autopilot~\cite{Autonomous2020}, immersive communication~\cite{ImmersiveCom2020}, computer animation \cite{parent2012computer}, and virtual/augmented reality~\cite{zenner2020immersive}. 
Despite of rapid development in 3D sensing technology~\cite{sensing2015}, it is still difficult and costly to acquire 3D point cloud sequences with high temporal resolution (HTR), which hinders to finely represent deformable 3D objects~\cite{wang2018high}. Instead of relying on hardware development, 
we consider computational methods to construct an HTR point cloud sequence from one with low temporal resolution (LTR), as illustrated in Fig.~\ref{fig:scene}. 

\begin{figure*}
\centering
\subcaptionbox{}{
\includegraphics[width=0.25\linewidth]{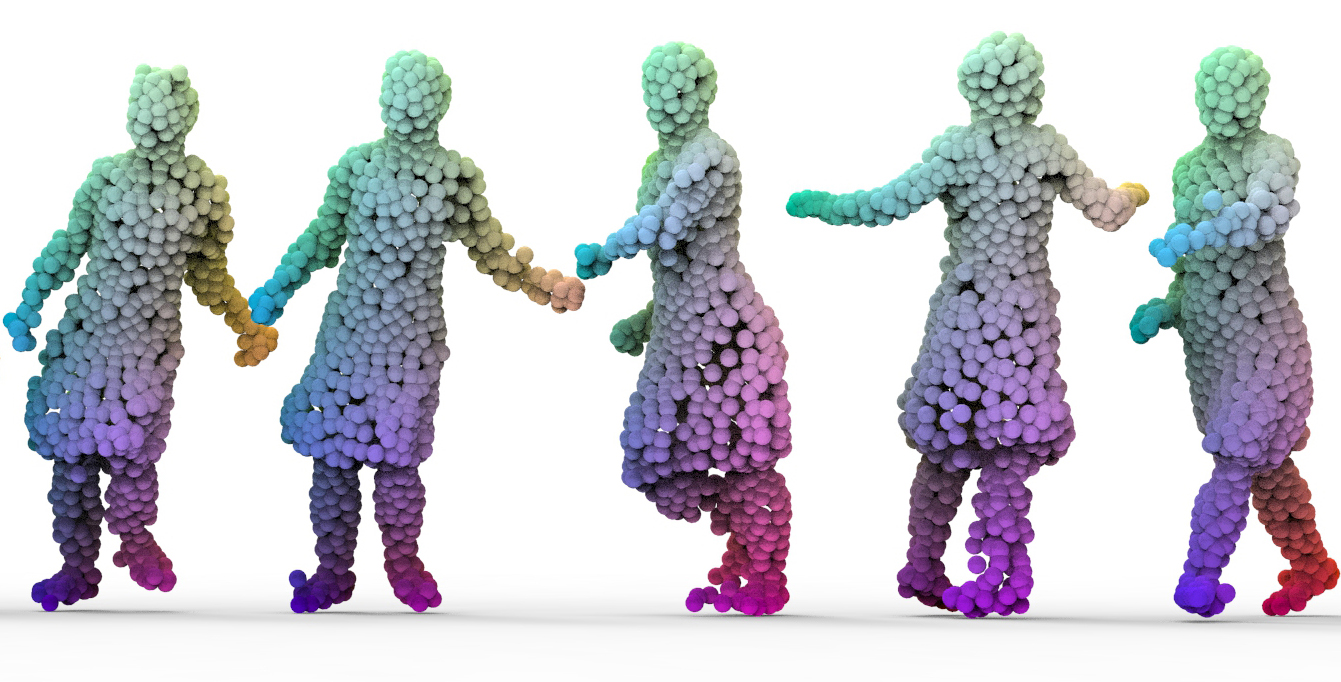}
}
\label{OurPINETswing_a} 
\subcaptionbox{}{
\includegraphics[width=0.25\linewidth]{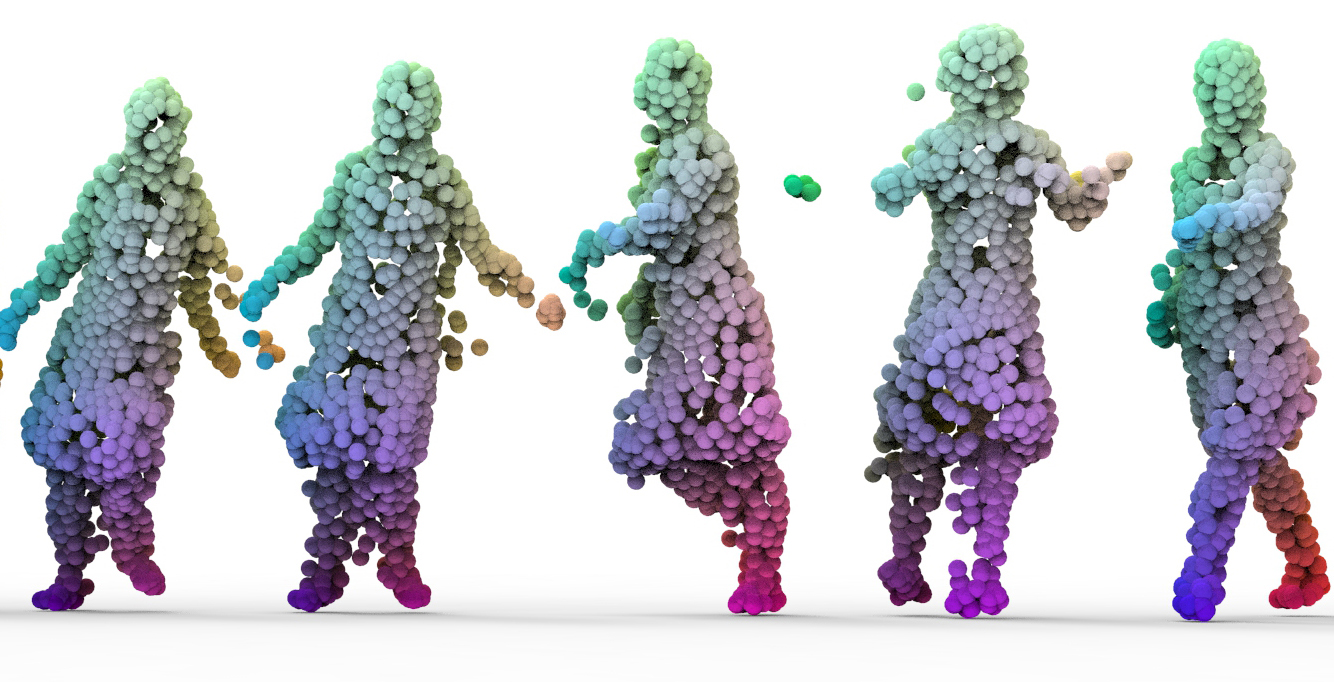}
}
\label{OurPINETswing_c} 
\subcaptionbox{}{
\includegraphics[width=0.25\linewidth]{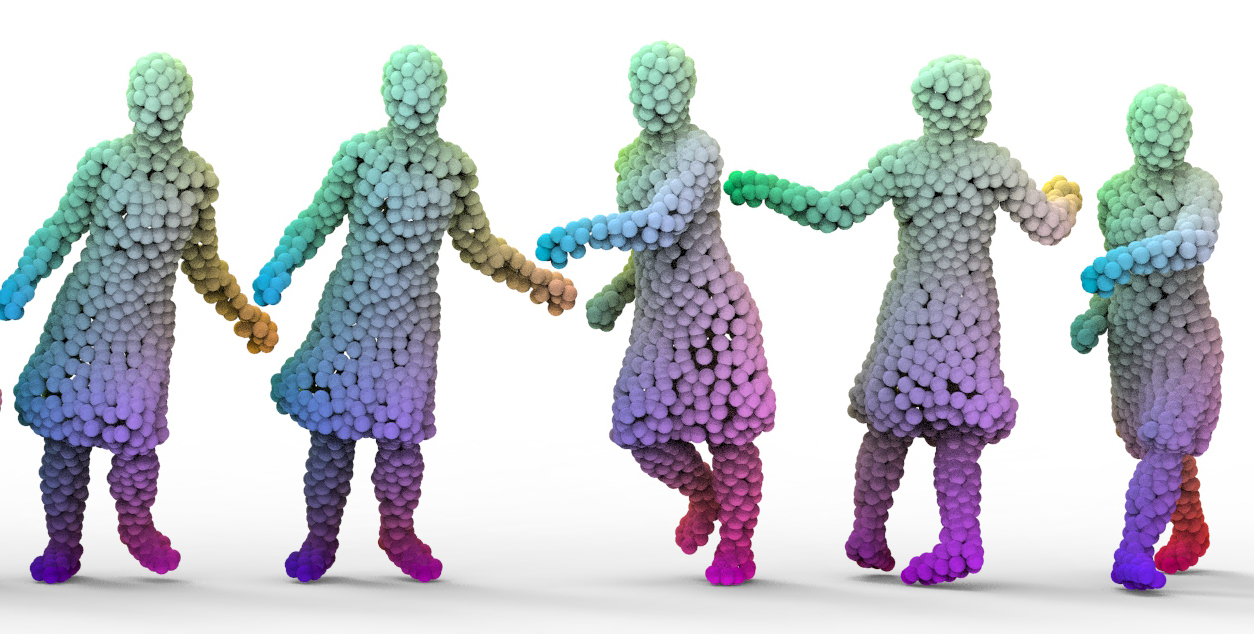}
}\vspace{-0.3cm}
\label{OurPINETswing_e} 
\caption{Visual comparisons of (a) our IDEA-Net, (b) PointINet \cite{lu2020pointinet}, and (c) the ground-truth on the \textit{Swing} sequence. 
}
\vspace{-0.4cm}
\label{fig_OurPINETswing} 
\end{figure*}

 
Although the considered problem 
shares similar properties with 2D video frame interpolation, both of which aim to interpolate/predict  the in-between frames of any two consecutive frames of an LTR sequence, the essentially different data modality (i.e., illumination vs. geometry information) 
makes it non-trivial to extend existing 2D video frame interpolation methods~\cite{jiang2018super,meyer2018phasenet,kalluri2021flavr} to 3D point clouds. Moreover, the unordered and irregular nature 
of 3D point cloud data in spatial and temporal domains poses great challenges.

Recently, several deep learning-based interpolation methods for 3D point cloud sequences 
have been proposed~\cite{qi2017pointnet,rakotosaona2020intrinsic,lu2020pointinet,groueix20183d}. 
Nevertheless, for the flow-based PointINet~\cite{lu2020pointinet}, 
it is mainly applicable to shapes with nearly rigid transformation 
and cannot well generalize to those with large non-rigid deformation. 
For the auto-encoder-based methods like~\cite{qi2017pointnet,rakotosaona2020intrinsic}, which directly interpolate  global features, 
since the
global features are abstract and insufficient to describe the details of motion changes, 
the interpolated frames tend to have similar shape appearances and lack temporal continuity, 
leading to stuck motion sequences. 
Besides, they are architecturally designed as separate learning stages, 
instead of fully end-to-end, which may suffer from severe information loss. 
Unlike existing works, 
we seek to build an interpretable interpolation framework with a clear geometric explanation. Moreover, in terms of application scenarios, we are interested in challenging dynamic point cloud data with large non-rigid deformation.

Technically,  we formulate the problem as estimation of point-wise trajectories (i.e., smooth curves in 3D Euclidean space) and reason that the  challenges are mainly posed by 
temporal irregularity and under-sampling,
which motivates us to disentangle the problem, leading to 
a two-step learning process: \textit{i)} coarse linear interpolation and \textit{ii)} trajectory compensation.  
Based on the explicit formulation, we propose IDEA-Net, an end-to-end deep interpolation framework, which features a dual-branch structure and consists of three steps: \textit{1)} extracting point-wise high-dimensional features, \textit{2)} learning point-wise temporal consistency and deducing coarse trajectories/in-between frames via linear interpolation, and \textit{3)} exploiting temporally regularized features to compensate the non-linear components of smooth trajectories. 
Experiments on both synthetic and real-scanned data demonstrate our IDEA-Net quantitatively and visually outperforms state-of-the-art methods to a large extent,  
as visualized in Fig. \ref{fig_OurPINETswing}. We also conduct extensive ablation studies to validate the rationality of our design.

In summary, we make the following contributions:
\begin{itemize}
    \item a new formulation for the problem of temporally interpolating dynamic 3D point cloud sequences; 
    \item a symmetric and coarse-to-fine network for end-to-end reconstructing HTR point cloud sequences from LTR point cloud sequences with large non-rigid deformation.
\end{itemize}
\section{Related Work}\label{related-work} 
\textbf{Deep 2D video frame interpolation}
aims to increase the frame rate of a video by generating the in-between frames. 
The existing methods can be generally classified into two categories: kernel-based and flow-based. The former 
~\cite{niklaus2017adaconv,niklaus2017sepconv,niklaus2018context,meyer2018phasenet} generates the in-between frames by convolution over local patches directly. 
The latter 
~\cite{xue2019video,jiang2018super,bao2019depth,liu2017video,liu2019deep} adopts the estimated flow to guide the warping process of the input frames. 
Different from 2D images/videos with the regular grid structure, 3D point clouds are characterized by both spatial and temporal irregularity, 
which impedes the direct extension of 
2D video frame interpolation models.

\textbf{Deep dynamic 3D point cloud processing}. 
The key challenges of this task lie in the temporal irregularity and large deformation in the point cloud sequence.
Existing techniques can be broadly classified into three types.
(1) Voxelize a point cloud sequence into a 4D volumetric grid \cite{luo2018fast,choy20194d,niemeyer2019occupancy}. 
For example, FaF~\cite{luo2018fast} uses 3D CNN to extract features. 
MinkowskiNet~\cite{choy20194d} analyzes the voxelized 4D tensor via a sparse 4D CNN.
(2) Adopt some sequential
 modules to deal with temporal information. 
For example, Yang \textit{et al.}~\cite{fan2019pointrnn} proposed PointRNN, PointGRU, and PointLSTM to model dynamic point clouds. 
(3) Directly perform sequence processing on raw points \cite{liu2019meteornet,prantl2019tranquil,rempe2020caspr,fan2020pstnet,lu2020pointinet,fan2021point,fan2021deep}. 
For example, \cite{liu2019meteornet,rempe2020caspr,fan2020pstnet,fan2021point}
perform feature aggregation by querying neighboring points in spatial and temporal domains. 
They have been applied to several tasks such as action recognition, pose estimation and segmentation. 
However, 
such aggregation is inaccurate, especially for point cloud sequences with large motions. 
To address this, 
PointINet~\cite{lu2020pointinet} adopts a scene flow estimator to interpolate two point clouds. 
However, PointINet fails to interpolate shapes with large deformation, e.g., human shapes.

\begin{figure*}[t]
	\centering
	\includegraphics[width=0.72\linewidth]{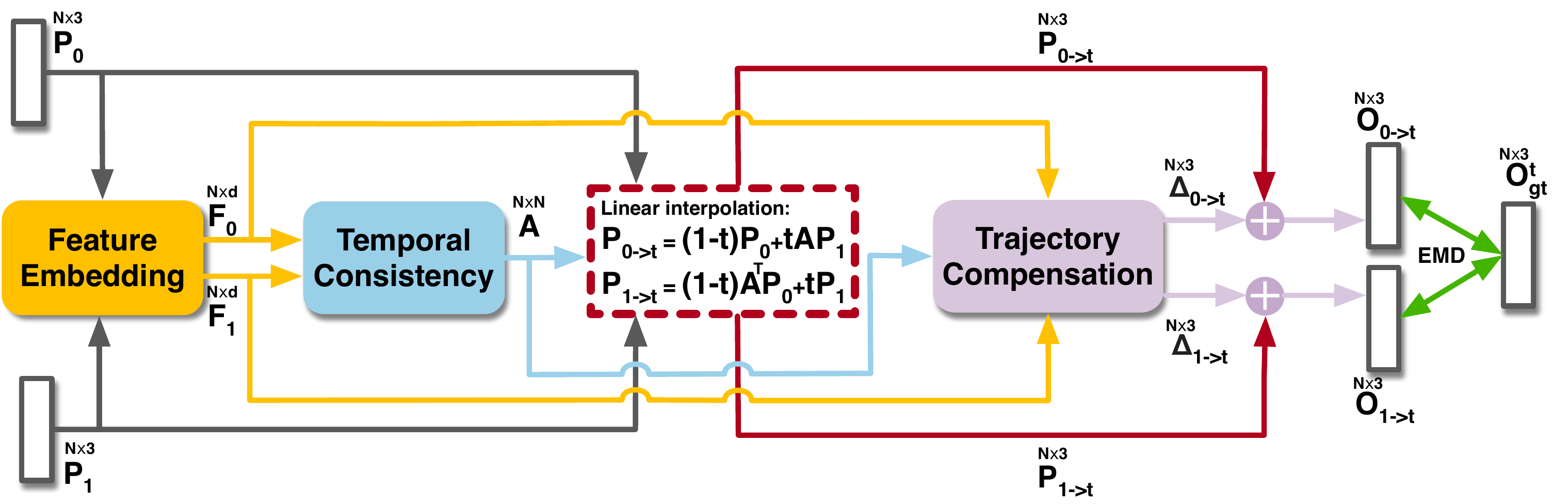}\vspace{-0.3cm}
	\caption{The flowchart of the proposed IDEA-Net for temporally interpolating any two consecutive frames of an LTR point cloud sequence in an end-to-end  manner. Besides, the user can vary the parameter $t$ in the range of $(0, 1)$ for interpolating frames continuously after training.  We refer the readers to \textit{Supplementary Material} for the detailed configuration of our network.}
	
	\label{fig:flowchart_all}
	\vspace{-0.5cm}
\end{figure*}

\textbf{Deep 3D shape interpolation}.
Inspired by 
the deep learning-based methods for 3D point cloud processing
~\cite{qi2017pointnet,qi2017pointnet++,wang2019dynamic,thomas2019kpconv}, 
a number of works adopting neural networks for 3D shape interpolation have 
been proposed~\cite{FoldingNet,achlioptas2018learning,groueix20183d,rakotosaona2020intrinsic,eisenberger2021neuromorph,lu2020pointinet},
which can be roughly divided into two categories.
 (1) Auto-Encoder (AE)-based methods.
For example, \cite{FoldingNet,achlioptas2018learning,groueix20183d} directly interpolate global features 
and feed the interpolated feature vector into the decoder to regress in-between point cloud frames.
\cite{rakotosaona2020intrinsic} further improves this kind of methods by introducing an edge AE trained with 3D meshes. 
However, interpolating global features without considering local region deformations may cause significant information loss. 
\cite{eisenberger2021neuromorph} adds normal information into the edgeConv \cite{wang2019dynamic} and calculates 
the geodesic matrices of remeshed 3D objects to explicitly constrain the learning of correspondence and shape interpolation for 3D meshes simultaneously. 
However, the required normals and topological information are unavailable in raw point clouds, making it non-trivial to extend the method to dynamic point cloud interpolation.
 Moreover, additional post-processing may be required to align the generated shapes or refine the correspondences~\cite{groueix20183d,rakotosaona2020intrinsic,eisenberger2021neuromorph}.
(2) Flow-based methods.
Similar to 2D image interpolation, \cite{lu2020pointinet} adopts the pre-trained flow estimation network for 3D point clouds, i.e., FlowNet3D~\cite{liu2019flownet3d}, 
to generate the bi-directional 3D flow, which is then used to warp the input frames to generate in-between estimations.
However, it cannot work well on data with large deformation due to the limitation of the adopted flow estimation.

\section{Proposed Method}\label{method}

\subsection{Problem Formulation} \label{subsec:formulation}\vspace{-0.5em}

\begin{figure}[t]
	\centering
    \includegraphics[width=0.7\linewidth]{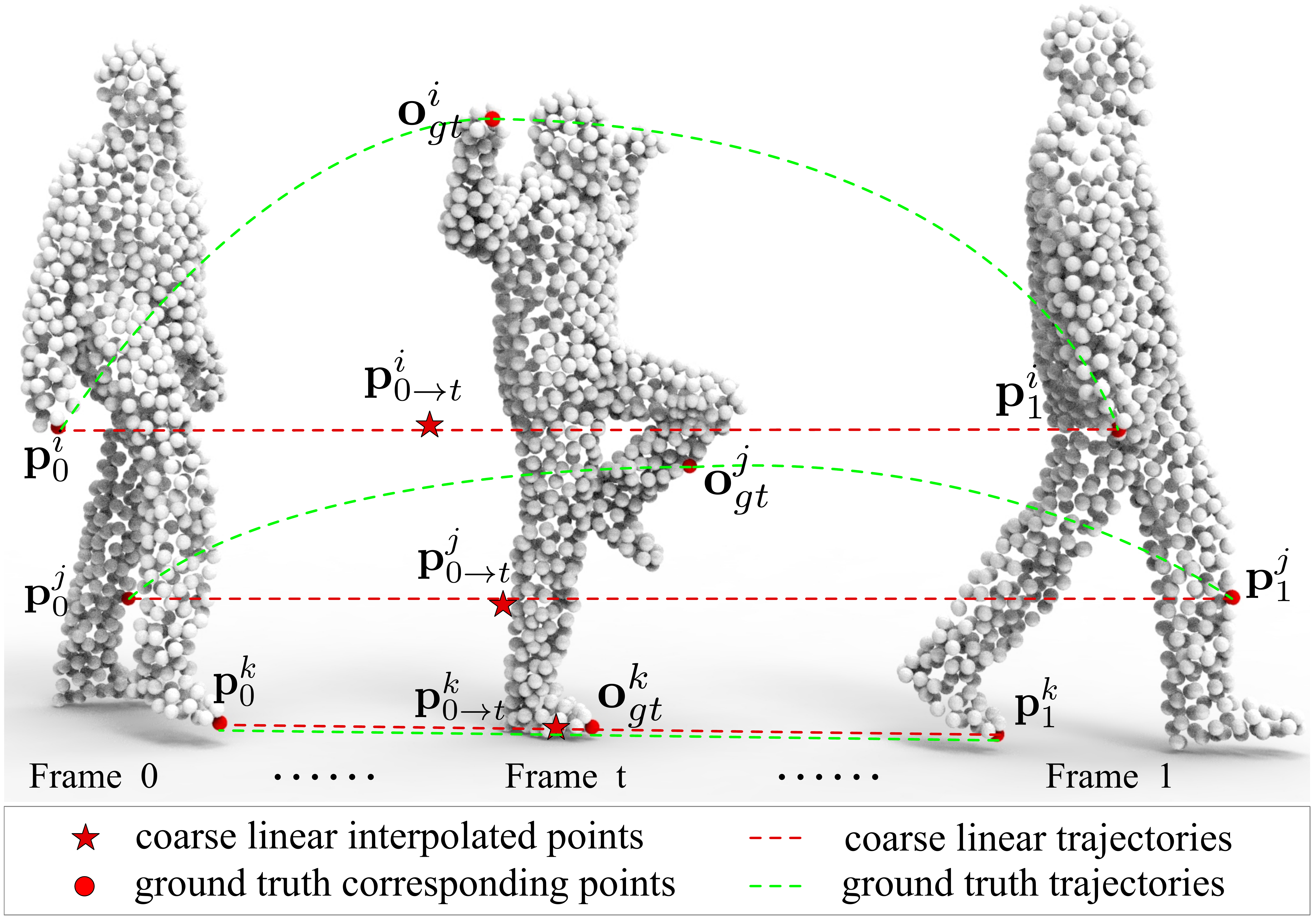}
    \vspace{-0.2cm}
	\caption{Illustration of the point-wise trajectories of typical points of a point cloud sequence,  where the \textcolor{green}{green} and \textcolor{red}{red} dashed lines denote the ground-truth and linearly interpolated trajectories, respectively.
	}
	\label{fig:tra_demo}
\end{figure}

Without loss of generality, let $\mathbf{P}_0\in\mathbb{R}^{N\times 3}$ and $\mathbf{P}_1\in\mathbb{R}^{N\times 3}$ be any two consecutive frames of an LTR point cloud sequence each with $N$ points\footnote{Note that the points contained in a point cloud frame are randomly stacked to form a matrix.}, and $\mathbf{p}_0^i$ and $\mathbf{p}_1^j\in\mathbb{R}^{1\times 3}$ the $i$-th and $j$-th points of $\mathbf{P}_0$ and $\mathbf{P}_1$, respectively. 
Assume that each point of $\mathbf{P}_0$ could be aligned to that of $\mathbf{P}_1$, and let the matrix $\mathbf{A}\in\mathbb{R}^{N\times N}$ explicitly encode such point-wise temporal consistency, 
i.e., 
if $\mathbf{p}_0^i$
corresponds to $\mathbf{p}_1^j$, 
$a_{ij}=1$; otherwise, $a_{ij}=0$. Let $\mathbf{a}_i\in\mathbb{R}^{1\times N}$ denote the $i$-th row of $\mathbf{A}$. Note that $\mathbf{A}$ is \textit{unknown}.

Obtaining an arbitrary point cloud frame between $\mathbf{P}_0$ and $\mathbf{P}_1$ 
 is equivalent to estimating the 
trajectory within each pair of points $\left\{\mathbf{p}_0^i,~ \mathbf{a}_i\mathbf{P}_1\right\}_{i=1}^N$. 
Generally, the trajectory of each point is a smooth curve in 3D Euclidean space; moreover, the fluctuation of the curves corresponding different points varies due to the articulated structure, non-rigid deformation, and other factors, as illustrated in Fig.~\ref{fig:tra_demo}. 
However, directly estimating such a curve only with two end-points may  have high uncertainty. 
 Thus, we disentangle this challenging problem and formulate it as a two-step coarse-to-fine process. Specifically,
we first uniformly approximate all point-wise trajectories by linear curve fitting, 
and accordingly the coarse in-between frame at time $\forall t\in (0,1)$ denoted as 
$\mathbf{P}_{0\rightarrow t}\in\mathbb{R}^{N\times 3}$ can be interpolated as 
\begin{equation}
\setlength{\abovedisplayskip}{3pt} 
\setlength{\belowdisplayskip}{3pt}
\mathbf{P}_{0\rightarrow t}
=(1-t)\mathbf{P}_0+t\mathbf{A}\mathbf{P}_1.
\label{equ:o_coarse}
\end{equation}
Although such a simple linear interpolation process is inaccurate, 
it is able to provide rational initialization to reduce ambiguity to some extent.
Then, to further compensate the high-order nonlinear components of trajectories missed in 
Eq.~(\ref{equ:o_coarse}) 
and correct the errors resulted from the inaccurate estimation of $\mathbf{A}$,
we introduce a trajectory compensation process.
Particularly, we can point-wisely map the input point clouds to a high-dimensional feature space by a typical nonlinear mapping function  $\phi(\cdot):\mathbb{R}^3\rightarrow\mathbb{R}^d$ and then estimate a 
function $f(\cdot):\mathbb{R}^d\rightarrow\mathbb{R}^d$ to
fuse the aligned features, which are finally transformed 
back to the point cloud space by another nonlinear function $\psi(\cdot):\mathbb{R}^d\rightarrow\mathbb{R}^3$ to obtain the increments ${\bf\Delta}_{0\rightarrow t}\in\mathbb{R}^{N\times 3}$ for trajectory compensation: 
\begin{equation}
\setlength{\abovedisplayskip}{3pt} 
\setlength{\belowdisplayskip}{3pt}
\mathbf{\bf\Delta}_{0\rightarrow t}= \psi\left( f(\phi(\mathbf{P}_0),\phi(\mathbf{P}_1),\mathbf{A},t )\right).
\label{equ:o_compensation}
\end{equation}
We expect that the high-order nonlinear components of the trajectories could be learned from the cues provided by the feature representation that can embed both local and global shape information of $\mathbf{P}_0$ and $\mathbf{P}_1$, as well as the contrast between the feature representations of them. 
The predicted in-between frame of an HTR sequence is finally obtained as 
\begin{equation}
\setlength{\abovedisplayskip}{3pt} 
\setlength{\belowdisplayskip}{3pt}
\mathbf{O}_{0\rightarrow t}=\mathbf{P}_{0\rightarrow t}+{\bf\Delta}_{0\rightarrow t}.
\label{equ:refine1}
\end{equation}

Moreover, because the transpose of the alignment matrix $\mathbf{A}^\textsf{T}\in\mathbb{R}^{N\times N}$ also depicts the temporal consistency information  
from $\mathbf{P}_1$ to $\mathbf{P}_0$, the previous formulation can be equivalently written as 
\begin{equation}\label{equ:rigid2}
\setlength{\abovedisplayskip}{3pt} 
\setlength{\belowdisplayskip}{3pt}
\mathbf{P}_{1\rightarrow t}=(1-t)\mathbf{A}^\textsf{T}\mathbf{P}_0+t\mathbf{P}_1,
\end{equation}
\begin{equation}\label{equ:o_nonrigid_dual}\vspace{-0.3cm}
\setlength{\abovedisplayskip}{3pt} 
\setlength{\belowdisplayskip}{3pt}
{\bf\Delta}_{1\rightarrow t}=\psi\left( f(\phi(\mathbf{P}_0),\phi(\mathbf{P}_1),\mathbf{A}^\textsf{T},1-t )\right),
\end{equation}
\begin{equation}\label{equ:refine2}
\mathbf{O}_{1\rightarrow t}=\mathbf{P}_{1\rightarrow t}+{\bf\Delta}_{1\rightarrow t}. 
\end{equation}
Ideally, the two point clouds represented by $\mathbf{O}_{0\rightarrow t}$ and  $\mathbf{O}_{1\rightarrow t}$ are the same. From 
Eqs. (\ref{equ:o_coarse})-(\ref{equ:refine2}),
it can be seen that the problems of dynamic point cloud interpolation mainly rely on 
the learning of the point-wise temporal consistency and the realization of the trajectory compensation process.

\subsection{Overview of our Framework}\label{subsec:overview} 
Based on the above formulation, we propose an end-to-end deep learning-based framework dubbed IDEA-Net, a \textit{dual-branch} network, which mimics the two equivalent coarse-to-fine processes. As shown in Fig.~\ref{fig:flowchart_all}, our IDEA-Net comprises of three modules: feature representation, learning point-wise temporal consistency, and 
trajectory compensation.
Specifically, 
the feature representation module first embeds 3D coordinates into a high-dimensional feature space by exploring both the local and global geometry of a point cloud, leading to the point-wise high-dimensional features. 
Taking 
the features as input, the temporal consistency module then predicts a relaxed matrix $\mathbf{A}$ with the alignment effect, which naturally induces 
the coarse interpolations
via Eqs. (\ref{equ:o_coarse}) and (\ref{equ:rigid2}). Finally, 
the trajectory compensation
module regresses the aligned and interpolated high-dimensional features with the learned $\mathbf{A}$ to generate nonlinear increments in a residual learning manner for compensating 
the high-order components of trajectories.
We empirically pick $\mathbf{O}_{0\rightarrow t}$ if $t<0.5$ and $\mathbf{O}_{1\rightarrow t}$, otherwise as the final interpolated frame.
In what follows, we will detail each module.

\subsection{Hierarchical Feature Representation} \label{sec:feature}
We employ DGCNN~\cite{wang2019dynamic} as our backbone 
to map 3D coordinates of input point clouds into a high-dimensional feature space, 
in which both local and global semantics are progressively embedded 
through the dynamic graph construction mechanism. Specifically, this module is composed of 
four layers of EdgeConv, which dynamically selects neighbours to aggregate local information to obtain point-wise features. 
Besides, a global feature that is formed by the adaptive max and average pooling for all point-wise features is concatenated to each local feature to get the final point-wise features, denoted as $\mathbf{F}_0\in\mathbb{R}^{N\times d}$ and $\mathbf{F}_1\in\mathbb{R}^{N\times d}$ for  $\mathbf{P}_0$ and  $\mathbf{P}_1$, respectively.  Denote by $\mathbf{f}_{0}^i$ and  $\mathbf{f}_{1}^j\in\mathbb{R}^{1\times d}$ the $i$-th and $j$-th rows of  $\mathbf{F}_0$ and $\mathbf{F}_1$, which encode the high-dimensional features of the $i$-th and $j$-th points of $\mathbf{P}_0$ and  $\mathbf{P}_1$, respectively. 
We refer the readers to~\cite{wang2019dynamic} for more details of DGCNN.

\subsection{Learning Point-Wise Temporal Consistency} \label{rigid}
As aforementioned, in reality, each frame of a point cloud sequence is captured individually in the camera space, 
resulting in temporal irregularity. 
Hence, we propose 
a temporal consistency module
to explicitly align the pair of input point clouds in point-wise, i.e., learning the matrix $\mathbf{A}$. 
However, the matrix $\mathbf{A}$ 
is ideally a 
binary permutation matrix, making it impossible 
to directly optimize it 
in a deep learning framework. To overcome the challenge, we optimize a relaxed alignment matrix, i.e.,  
$a_{ij}\geq 0$  
and $\mathbf{a}_i\mathbf{1}^{\textsf{T}}=1$ with $\mathbf{1}\in\mathbb{R}^{1\times N}$ being the vector whose all entries are one. Note that this module is end-to-end optimized in the IDEA-Net framework without additional 
supervision. 

Intuitively, a pair of aligned points 
should have similar semantic features. 
Motivated by this observation, we employ the distance between features to estimate $\mathbf{A}$.   Specifically, we first compute  
$\widetilde{\mathbf{A}}\in\mathbb{R}^{N\times N}$ whose $(i, j)$-th entry $\widetilde{a}_{ij}$ is 
\begin{equation}
\setlength{\abovedisplayskip}{3pt} 
\setlength{\belowdisplayskip}{3pt}
\widetilde{a}_{ij}=
1/\|\mathbf{f}_{0}^i-\mathbf{f}_{1}^j\|_2,
\end{equation}
where $\|\cdot\|_2$ returns the $\ell_2$-norm of a vector.
To further encourage $\mathbf{A}$ to mimic 
a binary matrix, we normalize the elements in each row
\begin{equation}
\setlength{\abovedisplayskip}{3pt} 
\setlength{\belowdisplayskip}{3pt}
	\widehat{a}_{ij}=(\widetilde{a}_{ij}-\mu_i)/\sigma_i,
\end{equation}
where $\mu_i$ and $\sigma_i$ are the mean and standard deviation of the $i$-th row of $\widetilde{\mathbf{A}}$.
Finally, we apply a softmax operator on $\widehat{\mathbf{A}}$ row-wisely to 
fulfill the relaxed constraint, generating
\begin{equation}
\setlength{\abovedisplayskip}{3pt} 
\setlength{\belowdisplayskip}{3pt}
a_{ij}=
e^{\widehat{a}_{ij}}/\sum_{j=1}^N{e^{\widehat{a}_{ij}}}.
\end{equation}
\textit{\textbf{Remark}.} Due to the non-differentiable characteristic of strictly binary matrices, our learned matrix $\mathbf{A}$ under such a relaxation process is no longer expected to 
exactly indicate point-wise temporal consistency relationships. 
In fact, as revealed in previous studies \cite{fan2021point}, 
since points may flow in and out across frames, there may not exist a ``ground truth'' point-wise consistency 
in most cases. 
Hence, we can interpret that $\mathbf{A}$ is functionally generalized to achieve coarse matching 
at both point and feature levels, and further drives the subsequent refinement module. 
Besides, the errors caused by the inaccurate estimation of $\mathbf{A}$ may be fixed in the subsequent refinement step to some extent.  
See Section \ref{sec:ablation} for the detailed ablation study on this module.

\subsection{Trajectory Compensation}\label{nonrigid}


With the learned 
$\mathbf{A}$ in Section \ref{rigid}, we can naturally obtain 
coarse interpolations via Eqs.~(\ref{equ:o_coarse}) and (\ref{equ:rigid2}).  
To simultaneously compensate the  nonlinear components of trajectories and fix 
the interpolation errors
caused by the inaccurate $\mathbf{A}$,
we introduce the trajectory compensation module.

Specifically, as the feature embedding process 
in Section \ref{sec:feature} can capture both the local and global geometric structures of the inputs, we employ it to realize the mapping function $\phi(\cdot)$ in Eq.~(\ref{equ:o_compensation}),  
i.e. 
\begin{equation}
\setlength{\abovedisplayskip}{3pt} 
\setlength{\belowdisplayskip}{3pt}
\mathbf{F}_0=\phi(\mathbf{P}_0),~ {\rm and}~~  \mathbf{F}_1=\phi(\mathbf{P}_1). 
\end{equation}
Considering that the 
non-linear trajectories could be deemed linear after being projected into the high-dimensional feature space via the data-driven manner, 
we simply implement the fusion function $f(\cdot)$ in Eq.~(\ref{equ:o_compensation}) with a linear function, i.e.,  
\begin{equation}
\setlength{\abovedisplayskip}{3pt} 
\setlength{\belowdisplayskip}{3pt}
\begin{split}
\mathbf{F}_{0\rightarrow t}&=(1-t)\mathbf{F}_0+t\mathbf{A}\mathbf{F}_1, \\
\mathbf{F}_{1\rightarrow t}&=(1-t)\mathbf{A}^\textsf{T}\mathbf{F}_0+t\mathbf{F}_1.
\end{split}
\end{equation}
Finally, we adopt a shared multi-layer perceptron (MLP) to implement the mapping $\psi(\cdot)$ for regressing  
the increments, i.e., 
\begin{equation}
\setlength{\abovedisplayskip}{3pt} 
\setlength{\belowdisplayskip}{3pt}
{\bf\Delta}_{0\rightarrow t} = \psi(\mathbf{F}_{0\rightarrow t}),~{\rm and}~~
{\bf\Delta}_{1\rightarrow t} = \psi(\mathbf{F}_{1\rightarrow t}).
\label{equ:increments}
\end{equation}
\textit{\textbf{Remark}}. Directly regressing an in-between point cloud frame via Eqs. (\ref{equ:o_compensation}) and (\ref{equ:o_nonrigid_dual}) 
seems to be a feasible solution. However, such an approach cannot yield satisfactory results (see the results in Section \ref{sec:ablation})
due to the curse of dimensionality. 
This also validates the necessity and rationality of the coarse linear interpolation module in 
Fig.~\ref{fig:flowchart_all}.

\subsection{Loss Function}
We train IDEA-Net end-to-end by simultaneously minimizing the earth mover's distance (EMD)~\cite{emdloss1998} between the reconstructed point cloud from each branch and the ground-truth one $\mathbf{O}_{gt}^t$: 
$$
\setlength{\abovedisplayskip}{3pt}
\setlength{\belowdisplayskip}{3pt}
\mathcal{L}=\frac{1}{2} \left( \mathcal{L}_{E M D}\left(\mathbf{O}_{0\rightarrow t}, \mathbf{O}_{gt}^t\right) + \mathcal{L}_{E M D}\left(\mathbf{O}_{1\rightarrow t}, \mathbf{O}_{gt}^t\right) \right)
$$
where $\mathcal{L}_{E M D}(\cdot, \cdot)$ computes the EMD between two point clouds.  

\textit{\textbf{Remark}}. Compared with the single-branch design, 
the proposed dual-branch design with shared network parameters reconstructs in-between frames from two directions, which is equivalent to regularizing $\mathbf{A}$ during training. 
Accordingly, the trained model can generate a more feasible $\mathbf{A}$ 
and like-wise better reconstruction quality during inference.   
Section~\ref{sec:ablation} demonstrates the superiority of such a dual-branch design via extensive evaluation.

\section{Experiments}\label{sec_exp}
\subsection{Experiment Settings} \label{sec_4_1}

\textbf{Datasets}.
We first constructed a dataset named  Dynamic Human Bodies dataset (DHB), containing 10 point cloud sequences from the MITAMA dataset~\cite{Vlasic2008ArticulatedMA} 
and 4 from the 8IVFB dataset\footnote{The MITAMA and 8IVFB datasets contain dynamic 3D meshes and real scanned point clouds, respectively. 
We uniformly sampled from each frame  1024 points.}. 
The sequences in DHB record 3D human motions 
with large and non-rigid deformation
in real world.
Besides, 
we adopted the commonly-used DFAUST\footnote{The DFAUST contains dynamic 3D meshes. Following the same setting as~\cite{rakotosaona2020intrinsic}, we uniformly sampled from each frame 1000 points.} dataset~\cite{dfaust_CVPR_2017}, a synthetic 3D human motion dataset used in Intrinsic Point Cloud Interpolation (PCI)~\cite{rakotosaona2020intrinsic} and PointNet AE~\cite{qi2017pointnet}, to fairly compare with them. 
For DHB, we used eight sequences to form the training set, and the remaining six sequences 
as the test set.  For DFAUST, following~\cite{rakotosaona2020intrinsic}  we used eleven action sequences to build the training dataset and three sequences for testing.  
We downsampled the acquired sequences in the temporal domain to generate input LTR point cloud sequences, i.e., we uniformly selected one frame every $k_{\text{train}}$ frames during the training phase and $k_{\text{test}}$ frames during the testing phase.  
See \textit{Supplementary Material} for more details of the datasets. 

\textbf{Compared methods}. 
We compared the most recent work named PointINet~\cite{lu2020pointinet}, which is flow-based and designed for temporal interpolation of point cloud sequences collected by LiDAR. For a fair comparison, we retrained its official code with the same DHB dataset as ours.
We also compared our method with two state-of-the-art AE-based methods, i.e., Intrinsic PCI~\cite{rakotosaona2020intrinsic} 
and PointNet AE~\cite{qi2017pointnet}. We adopted their pre-trained models on the DFAUST dataset
and trained our method with the same data as~\cite{rakotosaona2020intrinsic}.
It is also worth noting that Intrinsic PCI~\cite{rakotosaona2020intrinsic} requires the edges of a template mesh as extra input during training. Besides, we followed the setting of~\cite{rakotosaona2020intrinsic}, which adopts the ICP~\cite{chen1992object} as the post-processing, to align
the interpolated frames from Intrinsic PCI~\cite{rakotosaona2020intrinsic} and PointNet AE~\cite{qi2017pointnet} 
with the input frames.
Note that our method does not need a template mesh and any post-processing.

\textbf{Evaluation metrics}.\label{sec:metric} To quantitatively evaluate the interpolation quality, we provided both the average and the frame-by-frame Chamfer distance (CD) and EMD between the interpolated frames and the ground-truth ones of a sequence. Besides, we conducted subjective evaluation to compare different methods comprehensively. See Section~\ref{sec:subjective} for details.

\subsection{Experimental Results} \label{sec:result}

\begin{figure}[t]
\setlength{\abovecaptionskip}{0.1cm}
\setlength{\belowcaptionskip}{0.1cm}
\centering
\subcaptionbox{}{
\includegraphics[width=0.45\linewidth]{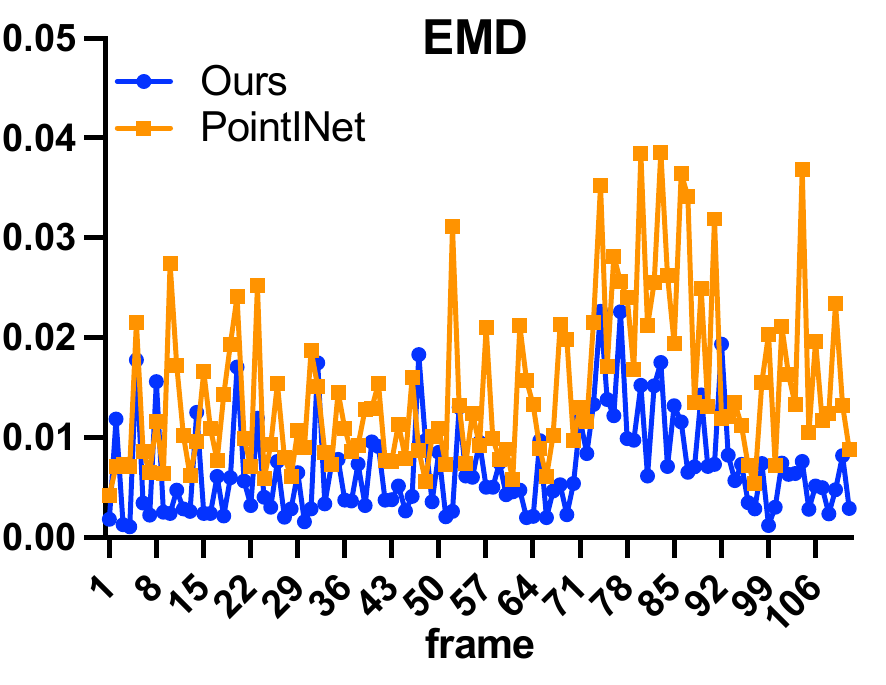}}
\label{_a} 
\subcaptionbox{}{
\includegraphics[width=0.45\linewidth]{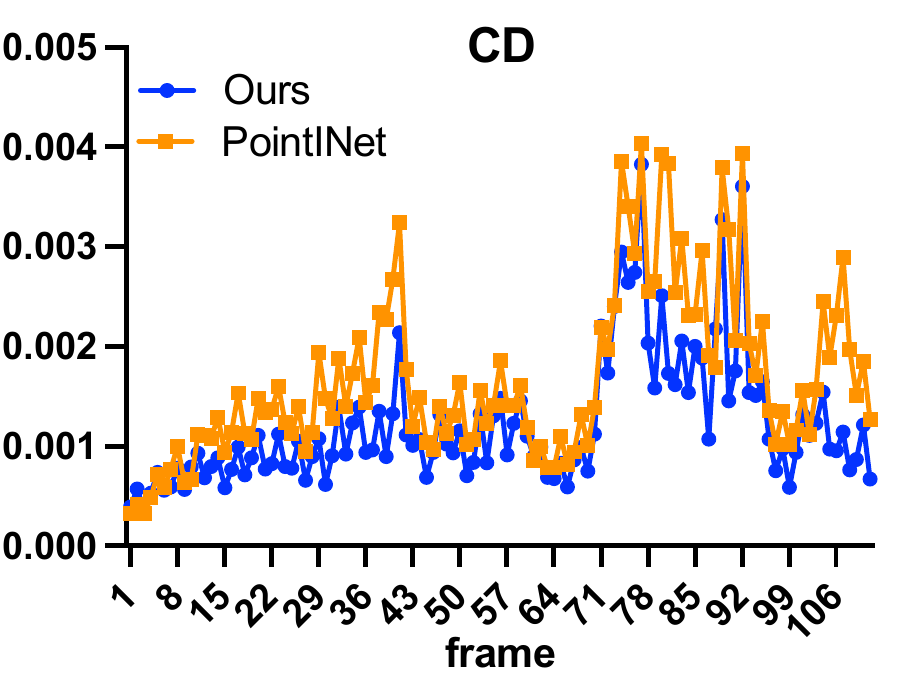}}
\label{_b} 
\vspace{-0.2cm}
\caption{Comparisons of the frame-by-frame quality of the reconstructed in-between frames on \textit{Swing} of the DHB dataset. 
}
\label{curve_1} 
\end{figure}

\begin{figure}[t]
\centering
\subcaptionbox{}{
\includegraphics[width=0.45\linewidth]{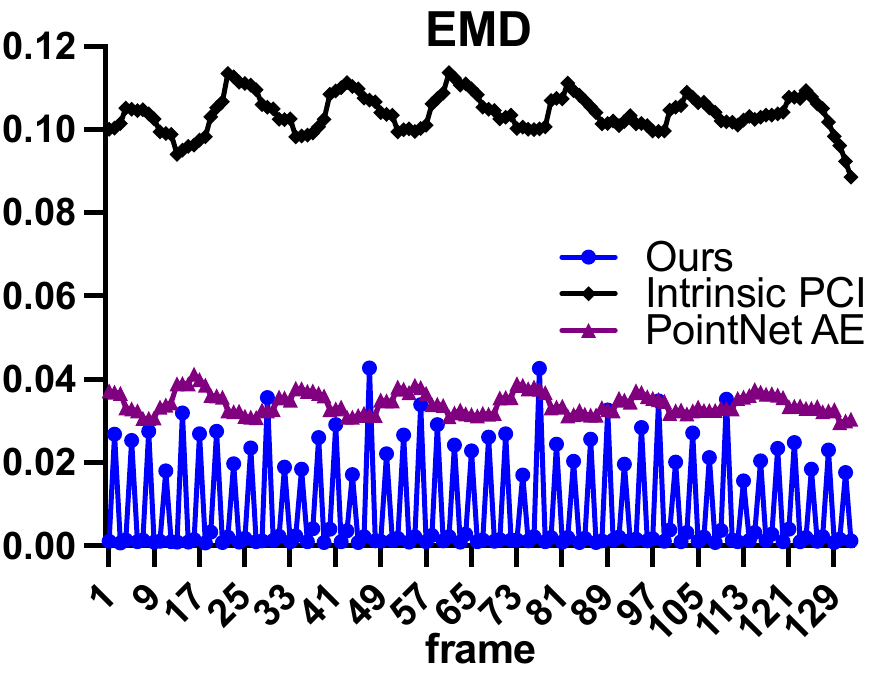}}
\label{__a} 
\subcaptionbox{}{
\includegraphics[width=0.45\linewidth]{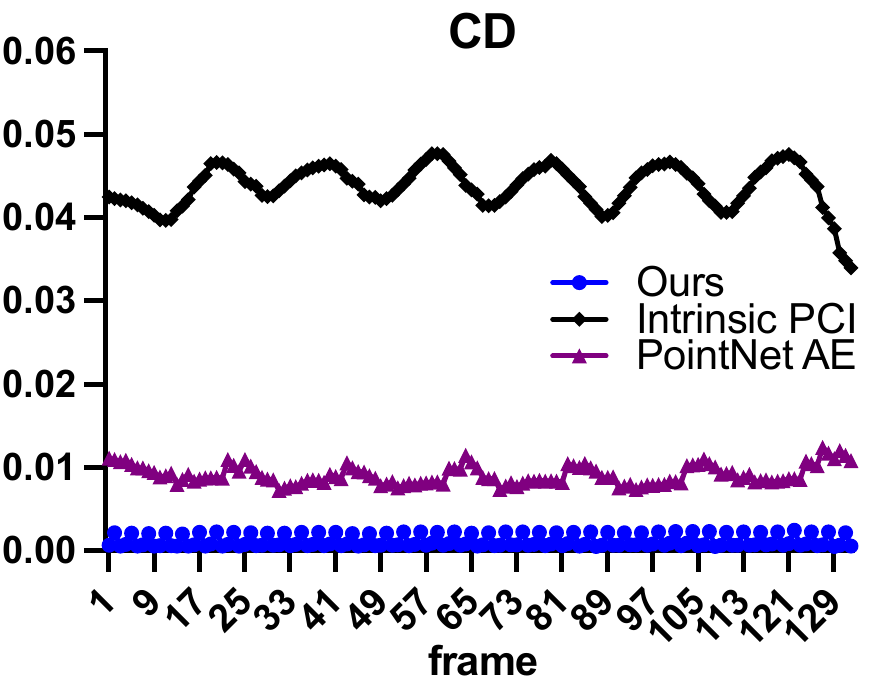}}
\label{__b} 
\vspace{-0.2cm}
\caption{
Comparisons of the frame-by-frame quality of the reconstructed in-between frames on \textit{shake\_arms} of the DFAUST dataset. 
}
\label{curve_2} 
\end{figure}

\begin{figure*}[t]
\centering
\subcaptionbox{}{
\includegraphics[width=0.3\linewidth]{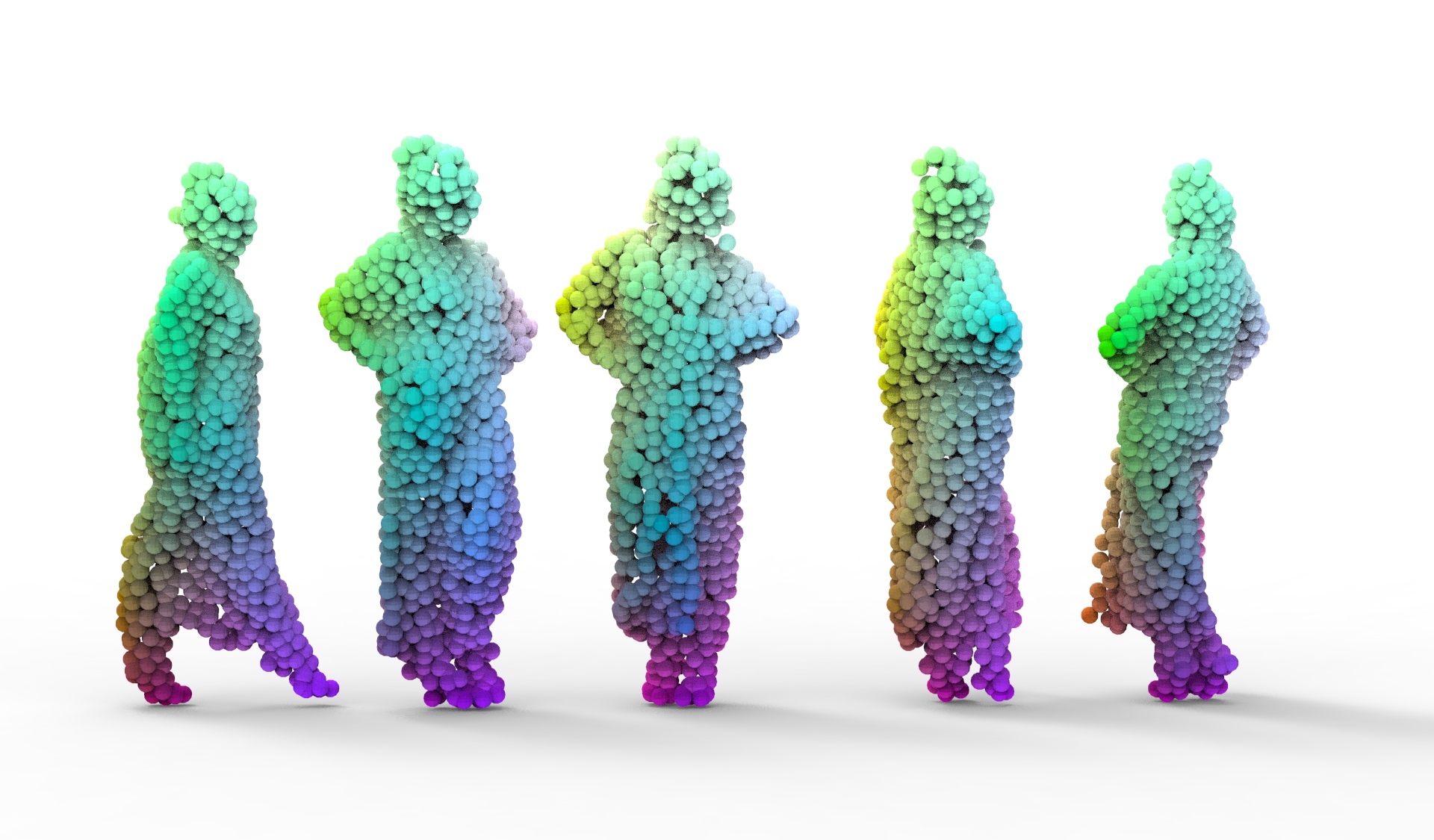}}
\label{OurPINETlongdress_a} 
\subcaptionbox{}{
\includegraphics[width=0.3\linewidth]{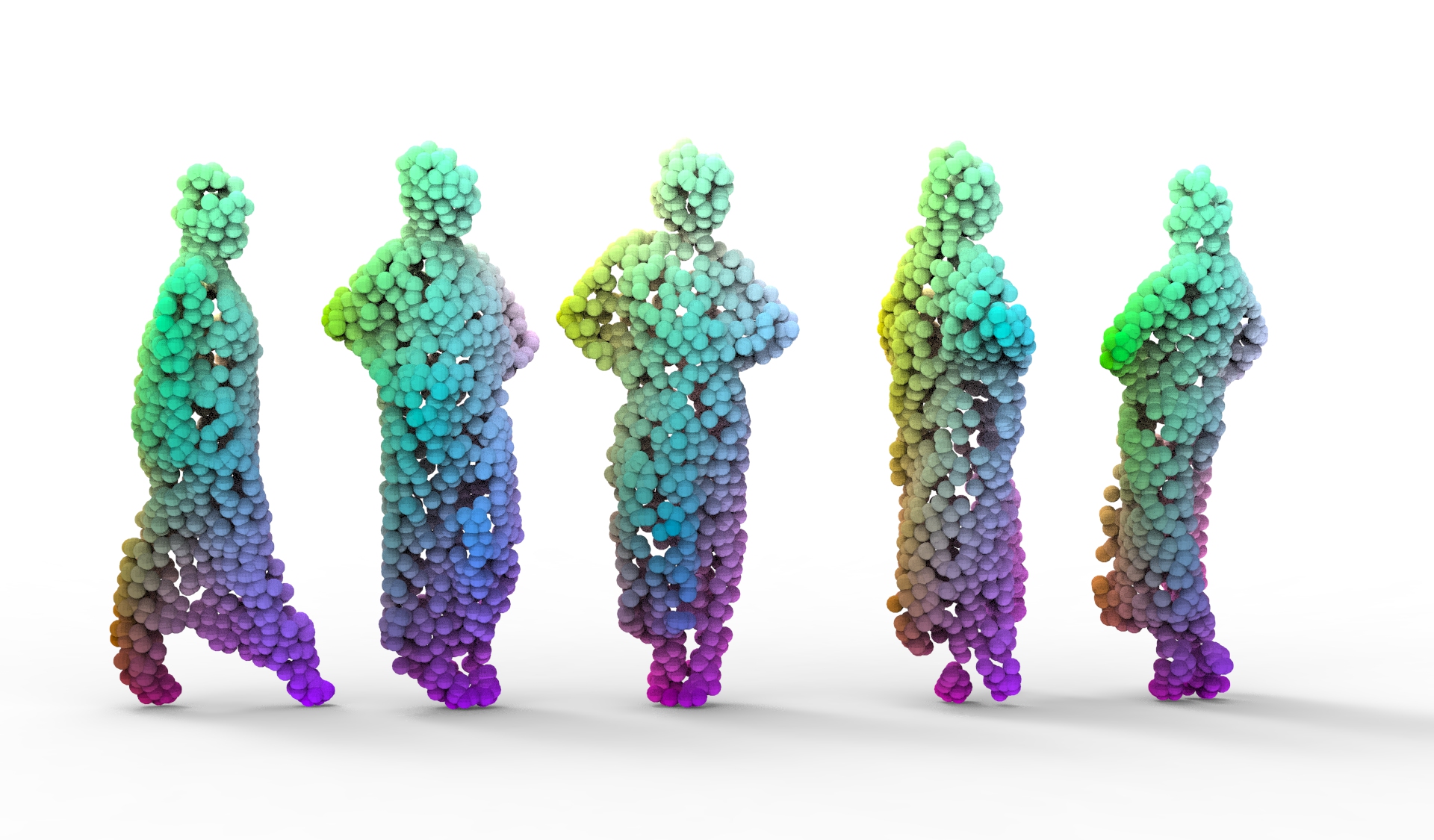}}
\label{OurPINETlongdress_b} 
\subcaptionbox{}{
\includegraphics[width=0.3\linewidth]{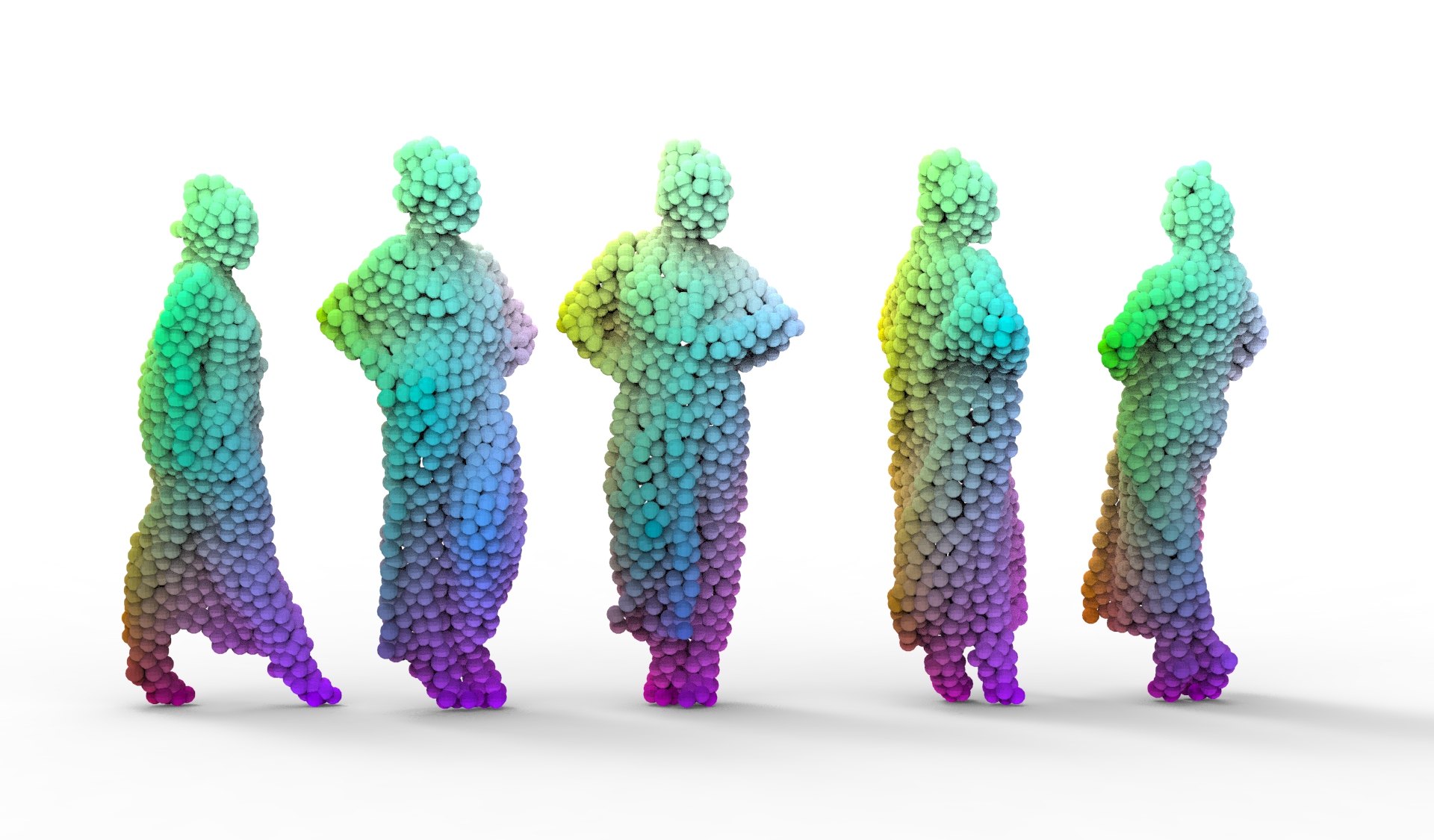}}
\label{OurPINETlongdress_c} 
\caption{Visual comparison of the interpolated in-between frames by   
(a) Ours and (b) PointINet ~\cite{lu2020pointinet}  on \textit{Longdress} of the DHB dataset, and   
(c) the corresponding ground-truth frames. 
\textit{Points of the interpolated frames by PointINet are non-uniformly distributed.}}
\label{fig_OurPINETlongdress} 
\end{figure*}

\begin{figure*}[t]
\centering
\subcaptionbox{}{
\includegraphics[width=0.43\linewidth]{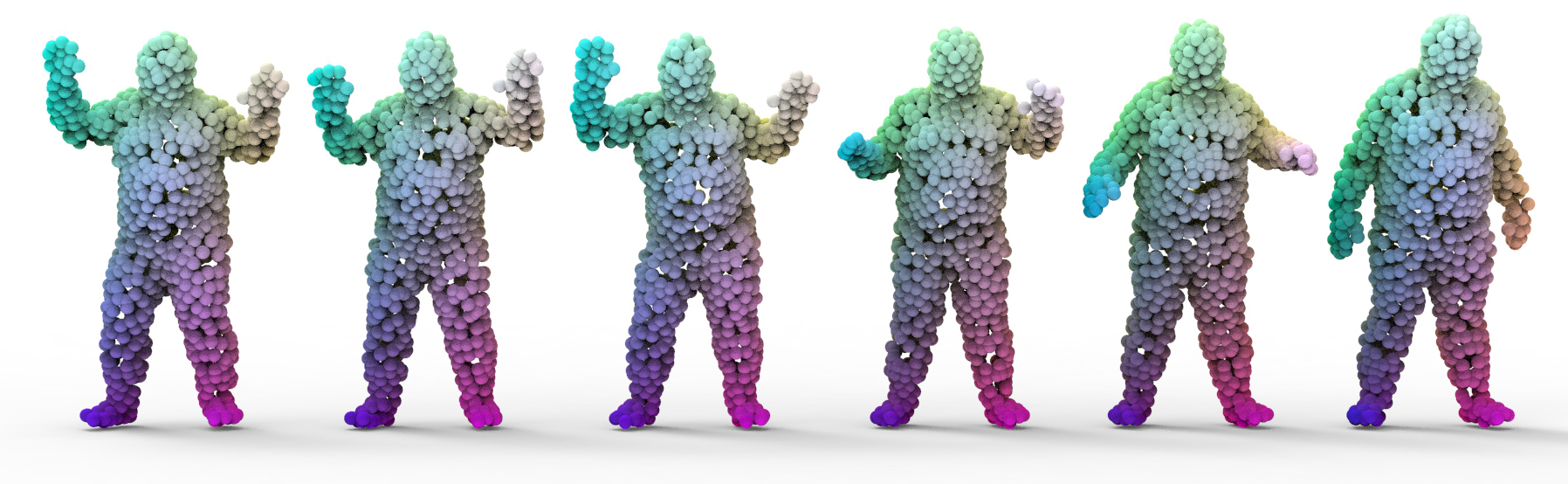}}
\label{OurECCV_a} 
\subcaptionbox{}{
\includegraphics[width=0.43\linewidth]{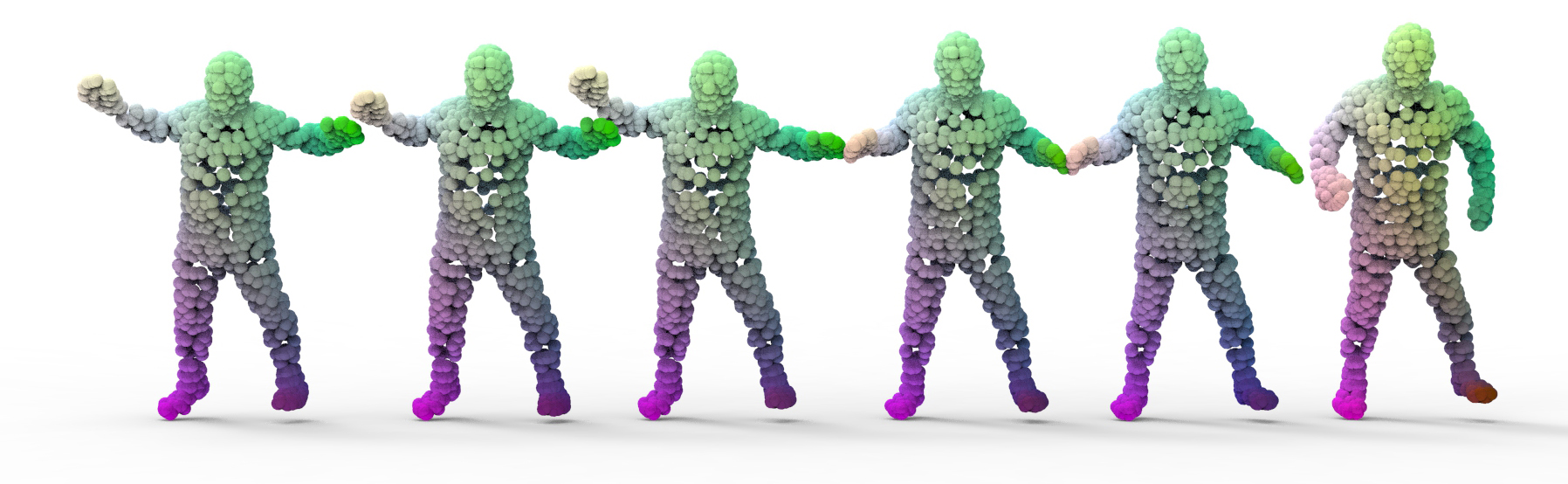}}
\label{OurECCV_c} 
\subcaptionbox{}{
\includegraphics[width=0.43\linewidth]{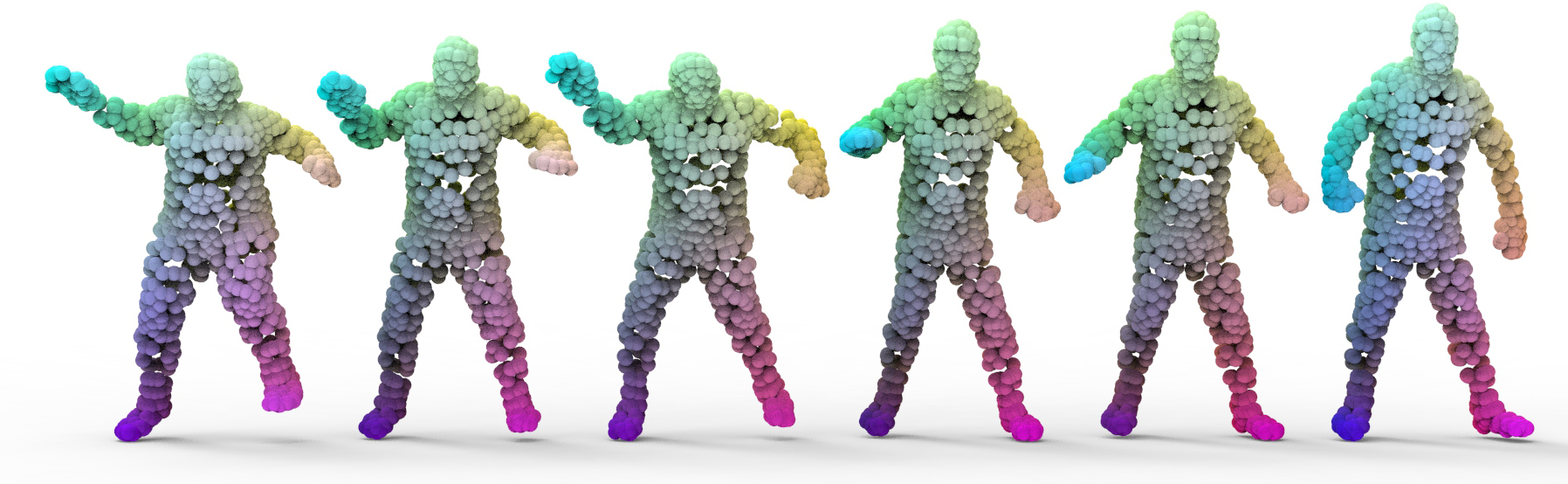}}
\label{OurECCV_e} 
\subcaptionbox{}{
\includegraphics[width=0.43\linewidth]{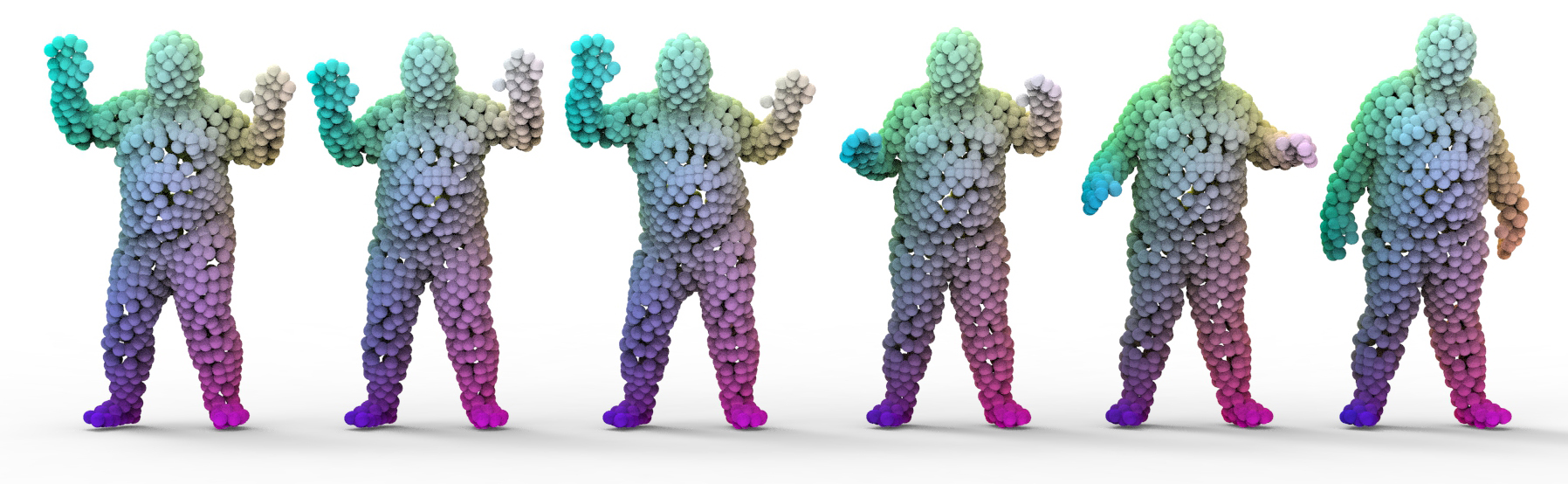}}
\label{OurECCV_g} 
\caption{Visual comparison of the interpolated in-between frames by (a) Ours,  (b) Intrinsic PCI~\cite{rakotosaona2020intrinsic}, and (c) PointNet AE~\cite{qi2017pointnet} on the sequence of the DFAUST dataset,  
and (d) the corresponding ground-truth frames. \textit{The overall shapes of the interpolated frames by Intrinsic and PointNet AE  obviously deviate from ground-truth ones.}
 }
\label{fig_OurECCV} 
\end{figure*}

\begin{table}[t]
\centering
\caption{\label{Tab_MITdyn_1} Quantitative ($\times10^{-3}$) comparison over the DHB dataset.  
}\vspace{-0.2cm}
\resizebox{0.3\textwidth}{!}{
\begin{tabular}{c|cc|cc}
\toprule 
 \multirow{2}{*} {Methods}
 & \multicolumn{2}{c|} {\text { Swing}} 
 & \multicolumn{2}{c} {\text { Longdress}} \\
\cline{2-5}
 \multicolumn{1}{c|} {\text {}} 
&\text{EMD} &\text{CD}
&\text{EMD} &\text{CD} \\
\midrule 
\text { PointINet ~\cite{lu2020pointinet} } &15.03 &1.70 & 10.09 &0.95\\
\text { Ours  } & $\mathbf{7.07}$ &$\mathbf{1.24}$& $\mathbf{5.92}$ &$\mathbf{0.88}$\\
\bottomrule
\end{tabular}
}
\vspace{0.18cm}
\end{table}

\begin{figure}[t]
\centering
\subcaptionbox{}{
\includegraphics[width=0.47\linewidth]{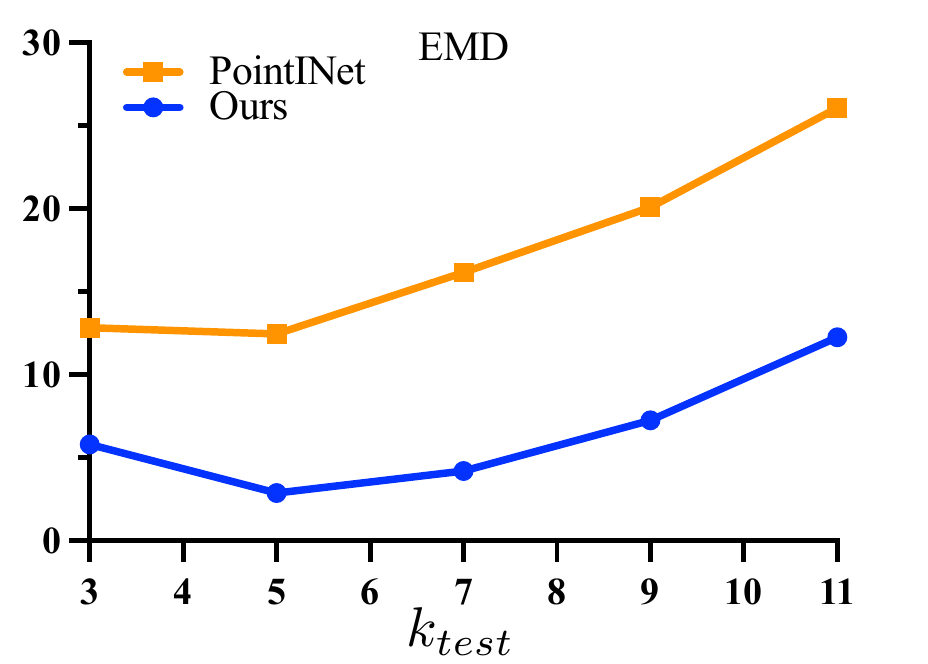}}
\label{6a} 
\subcaptionbox{}{
\includegraphics[width=0.47\linewidth]{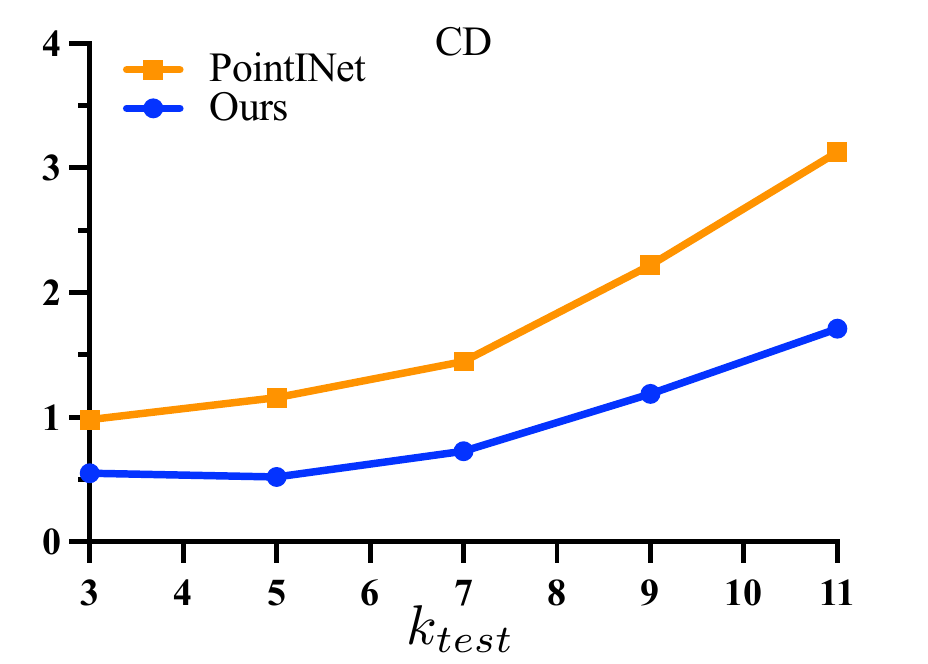}}
\label{6b} 
\label{6d} 
\caption{Flexibility verification 
on \textit{Squat\_2} of the DHB dataset. 
}
\label{fig8} 
\end{figure}
\begin{figure}[t]
\centering
\subcaptionbox{}{
\includegraphics[width=0.47\linewidth]{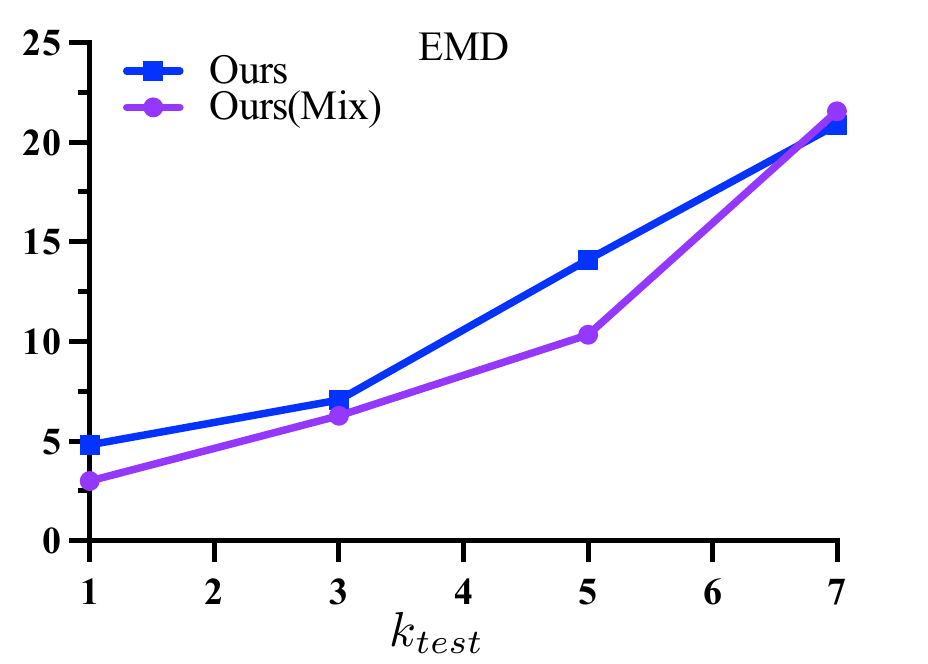}}
\label{7a} 
\subcaptionbox{}{
\includegraphics[width=0.47\linewidth]{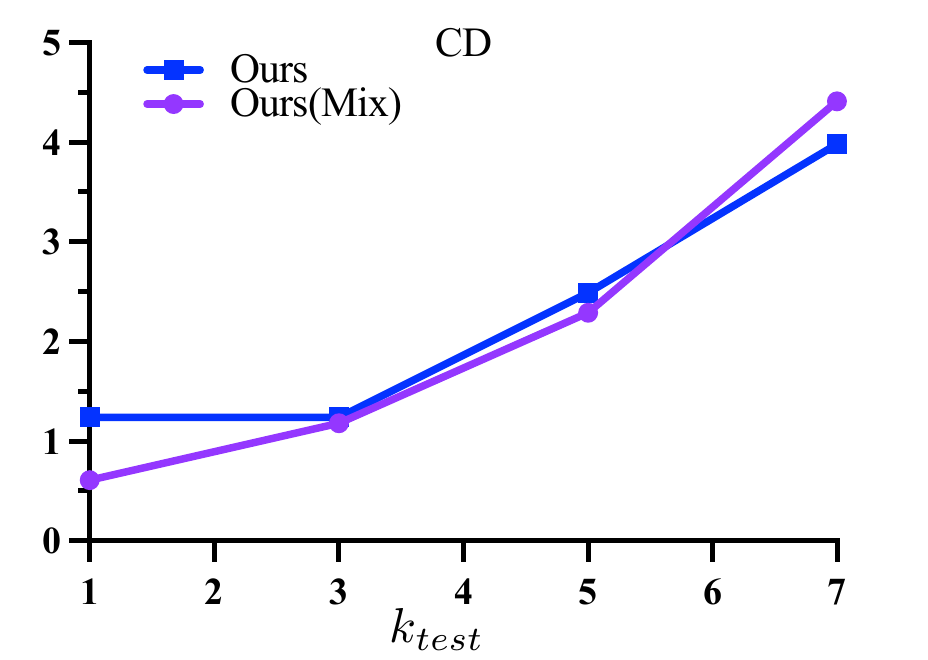}}
\label{7b}  
\caption{
Quantitative evaluation of our method trained with the mixed data training mechanism. The testing sequence is \textit{Swing} of the DHB dataset. 
}
\label{fig10} 
\end{figure}

\textbf{Results on the DHB dataset.} 
Table~\ref{Tab_MITdyn_1} shows the quantitative comparison with PointINet~\cite{lu2020pointinet}, where we set both $k_{\text{train}}$ and $k_{\text{test}}$  to 3 to generate LTR sequences for training and testing.
As Table~\ref{Tab_MITdyn_1} shows,
our method outperforms PointINet to a large extent under EMD metrics. 
The reason is that it is difficult for PointINet to explicitly predict accurate flows on sequences with large deformation, while our method is free of this operation. 
Besides, 
as shown in Fig.~\ref{curve_1}, 
our method achieves lower EMD and CD for \textit{most} frames 
and smaller fluctuations of frame-by-frame EMD and CD.
Fig.~\ref{fig_OurPINETswing} and Fig.~\ref{fig_OurPINETlongdress} show the visual comparison,  
where it can be seen that our method can generate 
frames that are closer to ground-truth ones, 
whereas PointINet~\cite{lu2020pointinet} tends to generate outliers and non-uniformly distributed points. 

\begin{table}[t]
\centering
\caption{\label{Tab_DFAust} Quantitative ($\times10^{-3}$) comparison over the DFAUST dataset. 
}\vspace{-0.2cm}
\resizebox{0.45\textwidth}{!}{
\begin{tabular}{c|cc|cc|cc}
\toprule
 \multirow{2}{*} {Methods} 
 & \multicolumn{2}{c|} {\text { shake\_arms}} 
 & \multicolumn{2}{c|} {\text { shake\_hips}} 
 & \multicolumn{2}{c} {\text { shake\_shoulders}}\\
 \cline{2-7}
 \multicolumn{1}{c|} {\text {}} 
&\text{EMD} &\text{CD} 
&\text{EMD} &\text{CD} 
&\text{EMD} &\text{CD}   \\
\midrule
\text { Intrinsic PCI ~\cite{rakotosaona2020intrinsic} } &103.85 &43.74  & 52.93 &22.04 &64.70 &28.55\\
\text { PointNet AE~\cite{qi2017pointnet} } & 34.23 & 9.09 & 40.85 & 12.35  & 29.18 & 11.99 \\
\text { Ours} & $\mathbf{9.31}$ &$\mathbf{1.20}$ &$\mathbf{6.85}$ &$\mathbf{0.91}$  &$\mathbf{9.19}$ &$\mathbf{0.95}$ \\
\bottomrule
\end{tabular}
}
\vspace{0.3cm}
\end{table}

\textbf{Results on the DFAUST dataset.} 
Table~\ref{Tab_DFAust} lists the quantitative comparison with Intrinsic PCI ~\cite{rakotosaona2020intrinsic} and PointNet AE~\cite{qi2017pointnet}, 
where we set both $k_{\text{train}}$ and $k_{\text{test}}$  to 3 to generate LTR sequences for training and testing. 
From Table~\ref{Tab_DFAust}, it can be seen that our method significantly outperforms Intrinsic and PointNet AE.  
The reason is that Intrinsic and PointNet AE adopt separate learning stages to 
vaguely interpolate global features for generating in-between frames, resulting in severe loss of spatial information, while our method is end-to-end and geometrically meaningful. 
From Fig.~\ref{curve_2}, 
it can be observed that our approach achieves much lower EMD and CD over almost all frames than Intrinsic and PointNet AE.
Fig.~\ref{fig_OurECCV} provides visual demonstration of our method,
from which we can see that these two AE-based methods fail to interpolate correct poses. Moreover, they cannot faithfully represent the original shapes.

\textbf{Evaluation of the flexibility.} 
To demonstrate the flexibility of our method,  we trained a single network with data that were generated by setting $k_{\text{train}}$ to 3 
and then evaluated the network with data that were generated by setting various $k_{\text{test}}\in \{3,5,7,9,11\}$. 
We also trained PointINet~\cite{lu2020pointinet} with the same setting for comparison. As shown in Fig.~\ref{fig8},  it can be seen that with the value of $k_{\text{test}}$ increasing, the interpolation problem becomes more challenging, and thus the reconstruction errors of both our method and PointINet gradually increase. However, our method always outperforms PointINet to a large extent, and especially the advantage is more obvious for a relatively larger $k_{\text{test}}$, demonstrating its stronger ability.

\textbf{Evaluation on the mixed data training mechanism.} 
In the previous experiments, we utilized all the ground-truth in-between frames for supervision. In this experiment, we set $k_{\text{train}}=4$ and $k_{\text{test}}\in \{1,3,5,7\}$ to generate training and testing data, respectively. During training, in each iteration we randomly selected only one of the in-between frames to-be-interpolated 
to optimize. Denote this training strategy as \textit{mixed data training}. 
Such a training manner can speed up the training process and improve the robustness of the network, owing to the diversity of data in different iterations.  
As shown in Fig.~\ref{fig10},  the performance of our method improves under this strategy,  
whereas PointINet~\cite{lu2020pointinet} suffers from serious performance degradation (see Table~\ref{Tab_mixpretrain}). This observation also demonstrates the advantage of our design.

\begin{table}[t]
\centering  
\caption{\label{Tab_mixpretrain}
Quantitative comparisons under  different training strategies.  (1)-(3): the methods were trained with data that were generated with $k_{\text{train}}=3$;  
(4)-(5): the methods were trained with the mixed data training strategy introduced in Section~\ref{sec:result}; Except for (3) where the flow estimation module of PointINet was pre-trained and fixed, the other networks were end-to-end trained from scratch. The testing data were generated by setting $k_{\text{test}}=3$.} 
\vspace{-0.2cm}
\resizebox{0.47\textwidth}{!}{
\begin{tabular}{l|cc|cc}
\toprule 
 \multirow{2}{*} {}
 & \multicolumn{2}{c|} {\text {Swing}} 
 & \multicolumn{2}{c} {\text {Longdress}} \\
\cline{2-5}
 \multicolumn{1}{c|} {\text {}} 
&\text{EMD} &\text{CD}
&\text{EMD} &\text{CD} \\
\midrule 
\text { (1) Ours } & 7.07 &1.24  & 5.92 &$\mathbf{0.88}$ \\
\text { (2) PointINet~\cite{lu2020pointinet}} &15.03 &1.70 & 10.09 &0.95 \\
\text { (3) PointINet~\cite{lu2020pointinet} (pretrained flow)} &15.38 &1.72 & 10.63 &0.96 \\
\text { (4) Ours (mixed) } & $\mathbf{6.74}$ &$\mathbf{1.21}$  & $\mathbf{5.84}$ &0.89 \\
\text { (5) PointINet~\cite{lu2020pointinet} (mixed) } &81.43 &14.19   & 84.49 &9.80 \\
\bottomrule
\end{tabular}
}
\vspace{0.3cm}
\end{table}

\subsection{Subjective Evaluation}\label{sec:subjective} 
\begin{figure}[t]
\centering
\setlength{\abovecaptionskip}{0.2cm}
\setlength{\belowcaptionskip}{0.2cm}
\subcaptionbox{\label{fig_subj_dhb}}{ 
\includegraphics[width=0.48\linewidth]{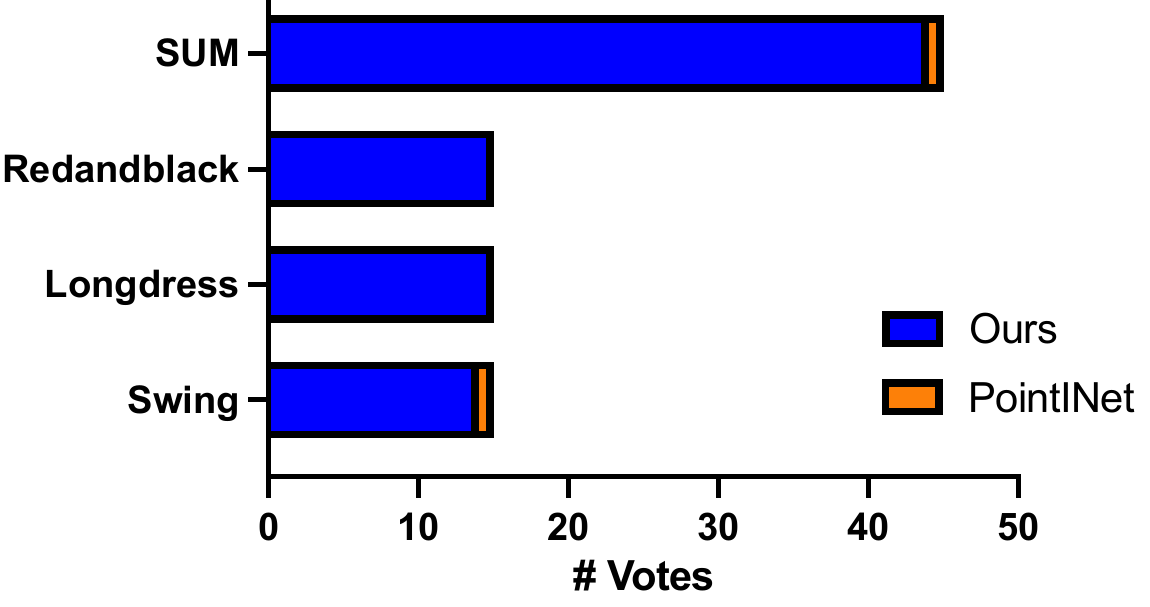}}
\subcaptionbox{\label{fig_subj_dfaust}}{ 
\includegraphics[width=0.48\linewidth]{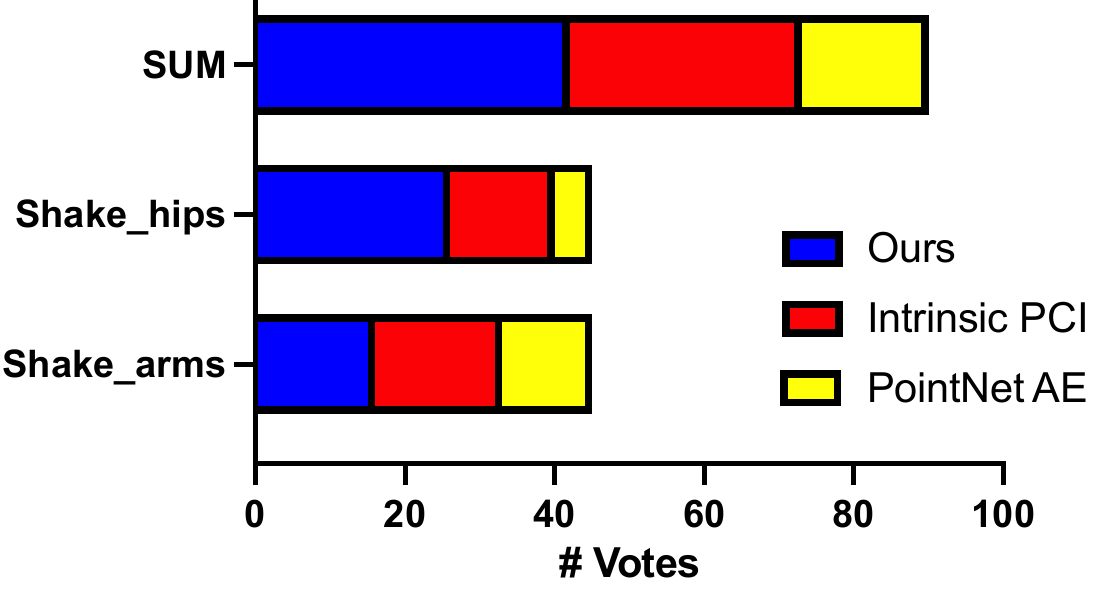}}
\caption{Subjective evaluation on the interpolated sequences by different methods on (a) DHB dataset and (b) DFAUST dataset.  ``SUM'' refers to the sum of votes on all testing sequences for each dataset.
}
\label{fig_subj} 
\end{figure}

To conduct the subjective evaluation, we displayed the interpolated sequences 
by all methods and  the corresponding ground-truth ones to 15 volunteers  
and asked them to vote for the method whose results they considered are closest to 
the ground-truth sequences. 
As shown in Fig.~\ref{fig_subj}, our IDEA-Net gets the highest number of votes on all the testing sequences, especially compared with the PointINet in Fig.~
\ref{fig_subj_dhb}. 
Besides, the evaluation results in  Fig.~
\ref{fig_subj_dfaust} indicate Intrinsic and PointNet-AE also obtain good subjective evaluations since these two methods can generate shapes with uniformly distributed points and few outliers. 
However, as illustrated in Fig.~\ref{fig_OurECCV}, these global feature-based methods fail to generate the correct poses and cannot preserve the faithful shapes.  
We also refer the readers to the Github page for video demos. 

\subsection{Ablation Study}\label{sec:ablation} \vspace{-0.4em}

\begin{figure}[t]
\centering
\subcaptionbox{\label{abl_a}}{
\includegraphics[width=0.23\linewidth]{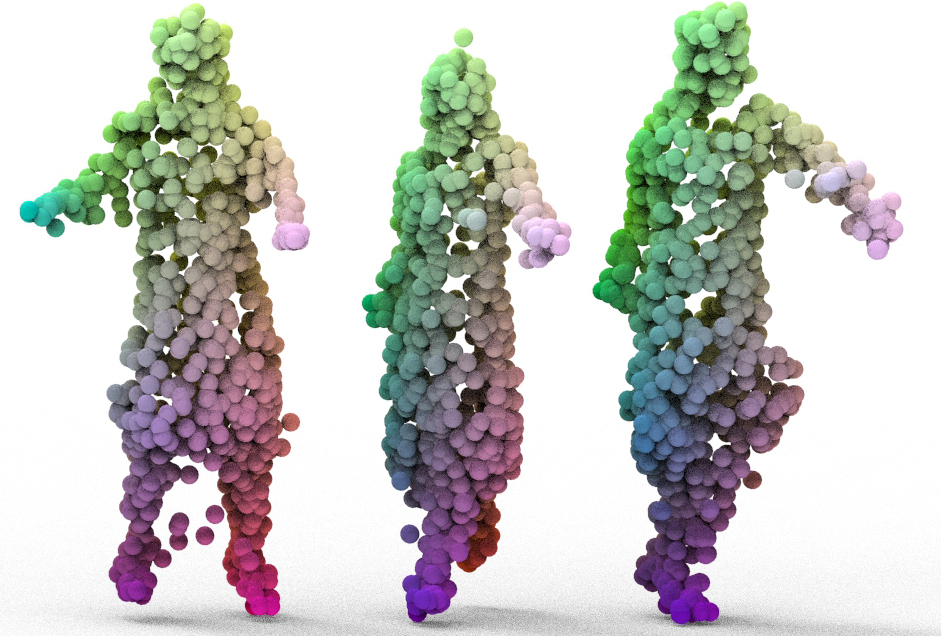}}
\subcaptionbox{\label{abl_b}}{
\includegraphics[width=0.23\linewidth]{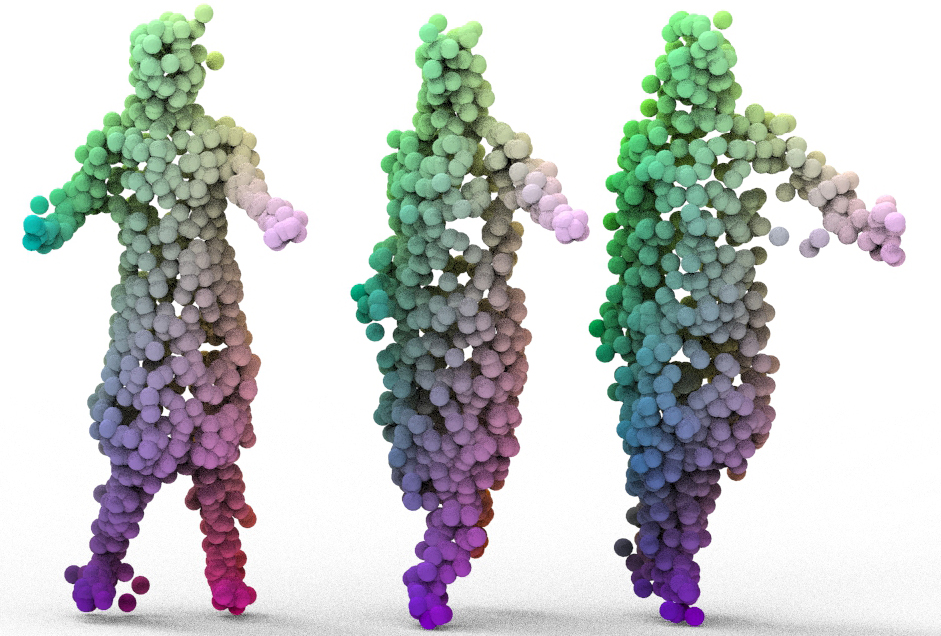}}
\subcaptionbox{\label{abl_c}}{
\includegraphics[width=0.23\linewidth]{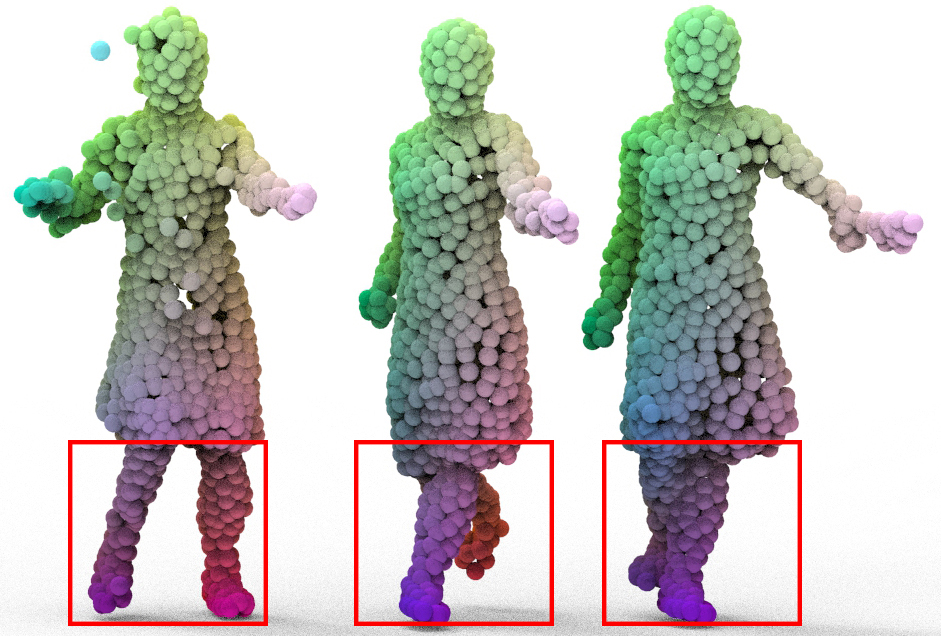}}
\subcaptionbox{\label{abl_f}}{
\includegraphics[width=0.23\linewidth]{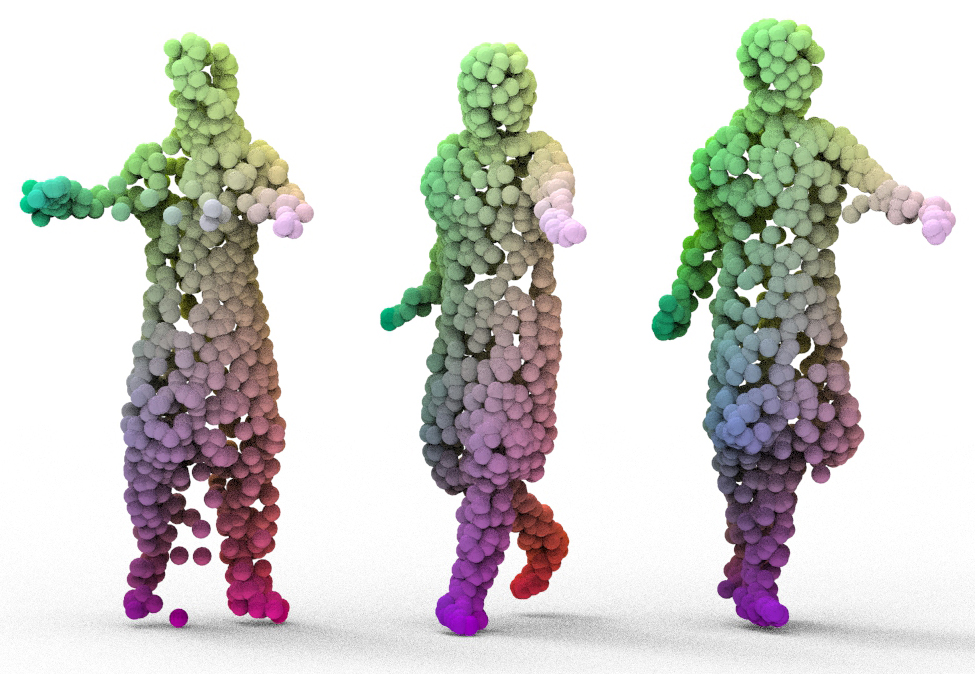}}\\
\subcaptionbox{\label{abl_d} }{
\includegraphics[width=0.23\linewidth]{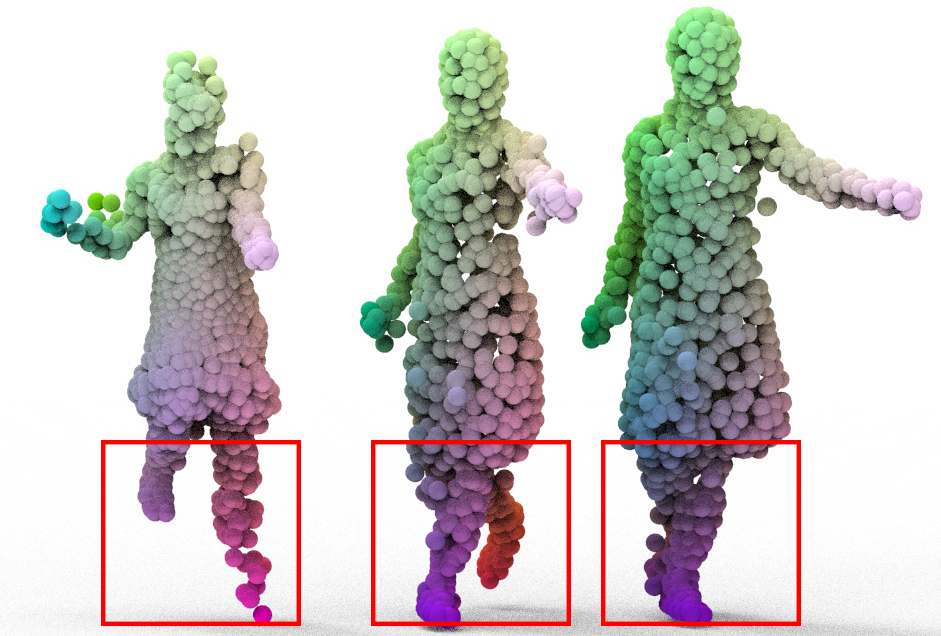}}
\subcaptionbox{\label{abl_e} }{
\includegraphics[width=0.23\linewidth]{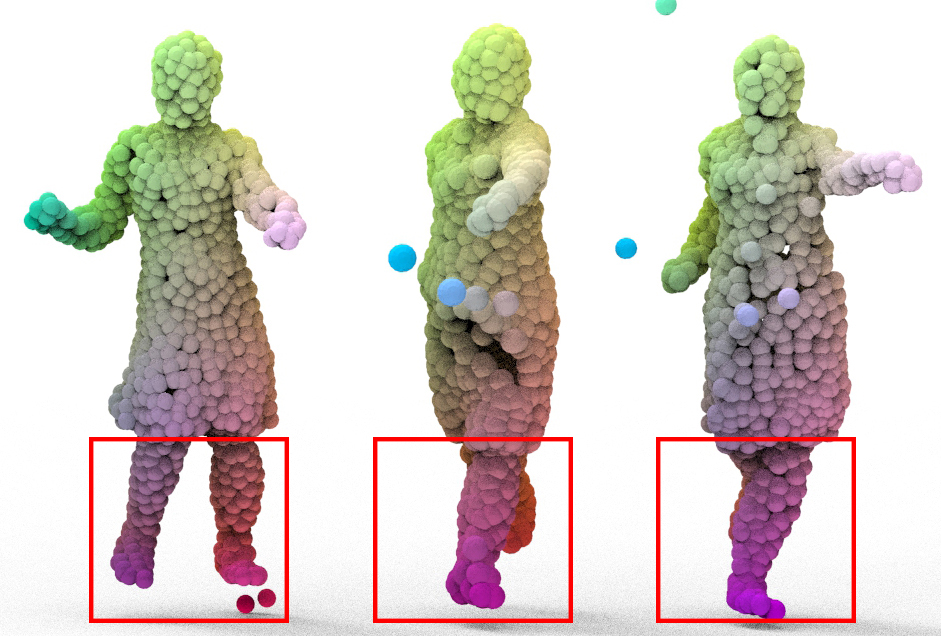}}
\subcaptionbox{\label{abl_our}}{
\includegraphics[width=0.23\linewidth]{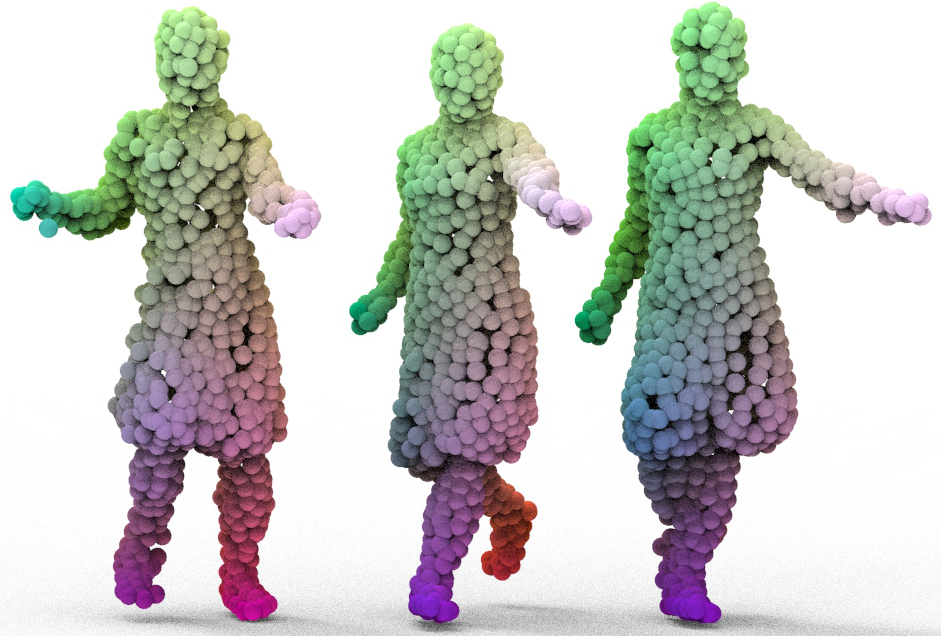}}
\subcaptionbox{\label{abl_gt}}{
\includegraphics[width=0.23\linewidth]{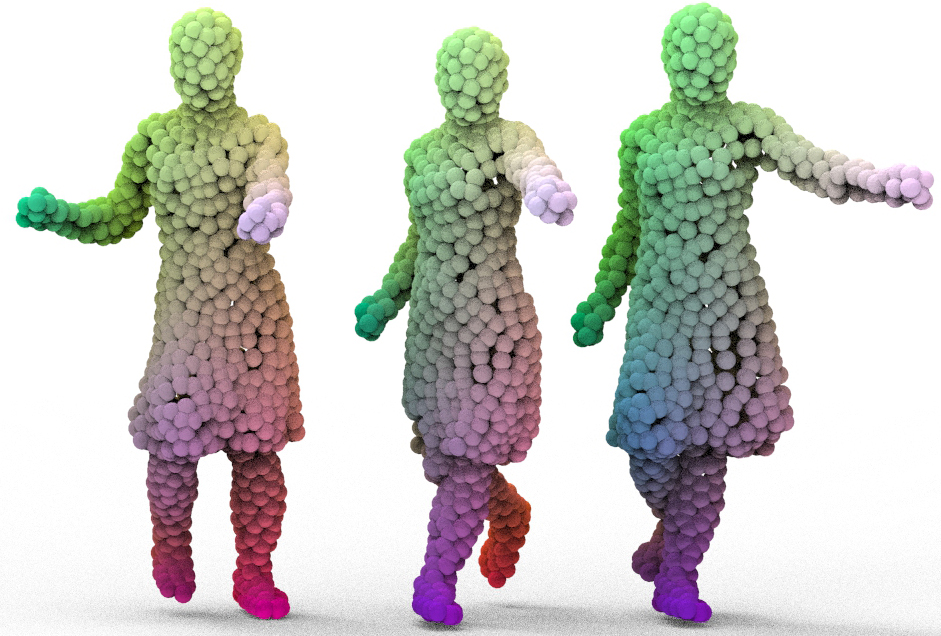}}
\caption{Visual comparisons of the results in ablation studies. (a)-(f) the interpolated frames by our method under the six ablation settings introduced in Section \ref{sec:ablation}, and (g) the proposed full model; (h) the ground-truth in-between frames. We highlighted the regions with serious shape distortion in \textcolor{red}{red} boxes.}
\label{fig_OurMixAblation} 
\vspace{-0.2cm}
\end{figure}

To comprehensively and deeply understand the effects of the core modules and architecture design of IDEA-Net, 
we conducted the following ablation studies. For each case, we retrained the modified network with the same training stragety as the full network and tested it on the same dataset.

\textbf{(a)-(b) 
Point-wise temporal consistency.}
We replaced this module with two types of 
matrices, i.e., 
a random permutation matrix and  a matrix $\mathbf{D}\in\mathbb{R}^{N\times N}$ with 
${d}_{ij}=1/\|\mathbf{p}_{0}^i-\mathbf{p}_{1}^j\|_2$.
As shown in Figs.
\ref{abl_a} and \ref{abl_b}, without this module, 
the resulting in-between frames show more wrong points, 
and the values of both EMD and CD increase significantly, demonstrating the effectiveness of this module.
We also refer the readers to \textit{Supplementary Material} for the visual illustration of learned $\mathbf{A}$.

 \textbf{(c) Linear interpolation.} 
We omitted the linear interpolation step and 
directly added $\mathbf{P}_0$ and $\mathbf{P}_1$ to the increments ${\bf \Delta}_{0\rightarrow t}$ and ${\bf \Delta}_{1\rightarrow t}$, respectively. 
We report the results in Table~\ref{Tab_ablation} (c) and Fig.~
\ref{abl_c}, 
showing that the loss gets worse and the generated shapes become less realistic than those of the full model, e.g., the shapes of the hands and legs.

\begin{table}[t]
\centering
\caption{\label{Tab_ablation}
Quantitative results ($\times10^{-3}$) of the ablation studies on the DHB dataset. 
(a)-(f) correspond to the six settings in Sec.~\ref{sec:ablation}.
}\vspace{-0.25cm}
\resizebox{0.3\textwidth}{!}{
\begin{tabular}{c|cc|cc}
\toprule 
 \multirow{2}{*} {Settings}
 & \multicolumn{2}{c|} {\text { Swing}} 
 & \multicolumn{2}{c} {\text { Longdress}} \\
\cline{2-5}
 \multicolumn{1}{c|} {\text {}} 
&\text{EMD} &\text{CD}
&\text{EMD} &\text{CD} \\
\midrule 
\text {Full model} & $\mathbf{6.74}$ &$\mathbf{1.21}$  & $\mathbf{5.84}$ &$\mathbf{0.89}$ \\
\hline
\text { (a) } &23.46 &3.11 & 17.82 &2.22 \\
\text { (b) } &25.02 &3.26   & 21.50 &2.26 \\
\text { (c) } &10.36 &2.38  & 6.33 &0.97 \\
\text { (d) } &18.28 &2.85  & 24.32 &2.57\\ 
\text { (e) } &11.00 &1.78  & 6.13 &1.40 \\
\text { (f) } &25.47 &9.05  & 29.85 &5.51\\
\bottomrule
\end{tabular}}\vspace{0.18cm}
\end{table}

 \textbf{(d) 
 Directly regress in-between frames from features.} Without employing the linear interpolation to generate coarse estimation, 
we 
directly regressed in-between point clouds from the interpolated feature derived by Eq. \eqref{equ:increments}. 
Table~\ref{Tab_ablation} (d) and Fig.~
\ref{abl_f} provide the quantitative and visual results, respectively, where it can be seen that  
the CD and EMD values increase more than twice,
and the resulting point clouds have considerable artifacts.

\textbf{(e) 
Trajectory compensation}. 
In Table~\ref{Tab_ablation} (e) and Fig.~
\ref{abl_d}, 
we reported the 
results of IDEA-Net without the compensation module, where it can be seen that without this module, the values of both EMD and CD increase and the resulting shapes are partially collapsed.

\begin{figure}[t]
    \centering
    \setlength{\abovecaptionskip}{0.2cm}
	\setlength{\belowcaptionskip}{0.2cm}
    \includegraphics[width=0.34\textwidth]{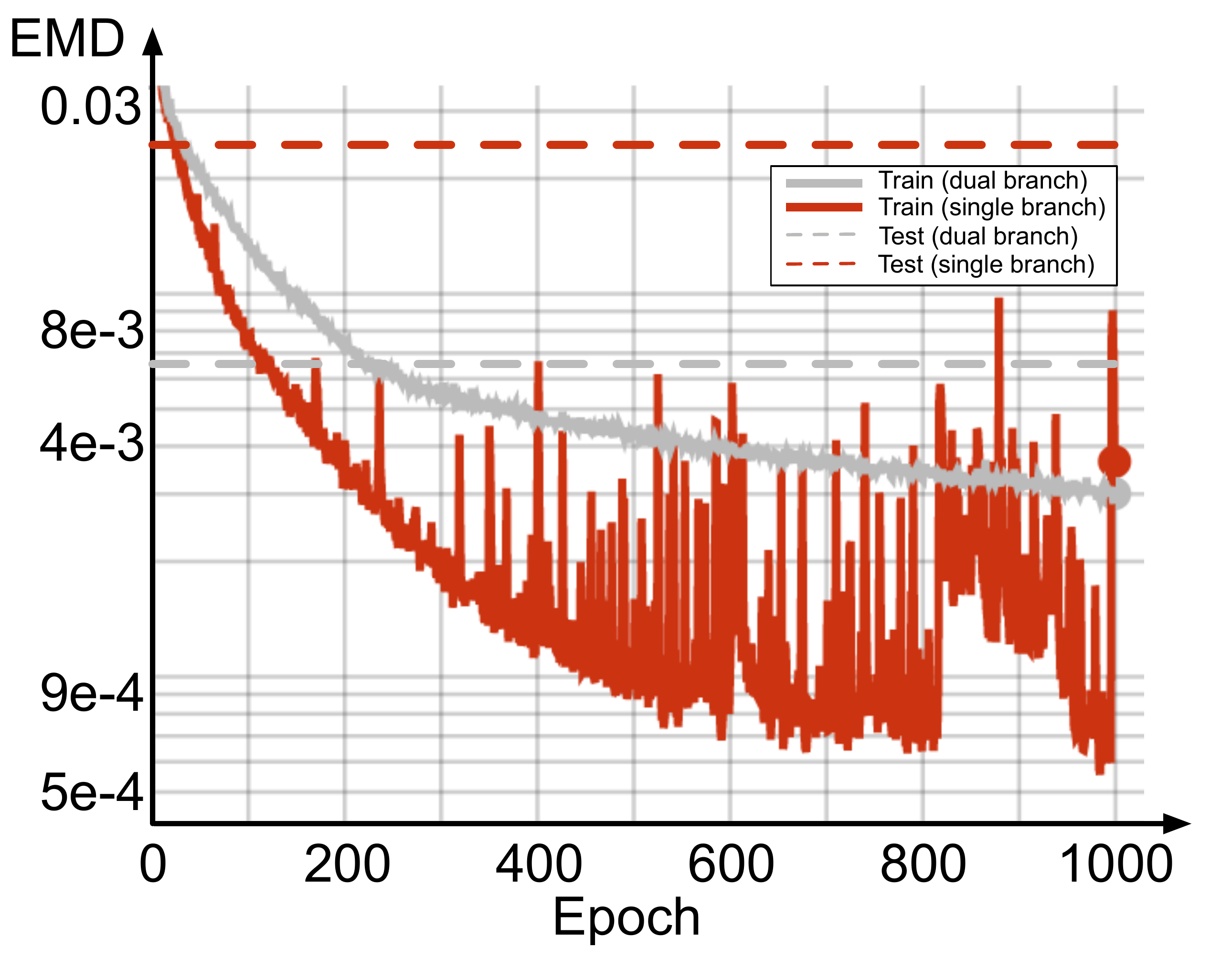}
        \caption{Comparison of the training and testing losses of the dual- and single-branch designs.}
    \label{fig:train-test-loss}
\end{figure}

\textbf{(f) 
Dual-branch design.} 
To conduct comparisons, we trained a single-branch network, i.e., the branch predicting $\mathbf{O}_{0\rightarrow t}$ in IDEA-Net.
As shown in Table~\ref{Tab_ablation} (f) and Fig.~
\ref{abl_e}, the single-branch model produces much larger EMD and CD than the dual-branch model (i.e., the full model). Besides, its visual results are noisy and the shapes (e.g., the legs) deviate from the ground-truth ones, validating the significant advantages of the symmetric dual-branch design with shared modules 
in terms of the reconstruction quality. 
Moreover, we showed the training (solid lines) and testing (dashed lines) losses for the dual- and single-branch models in Fig. \ref{fig:train-test-loss}. 
As can be seen, the dual-branch model can stabilize the training process. Also, it potentially avoids over-fitting and improves the generalization ability of the model, 
i.e., although the training losses of these two designs converge to comparable values, the testing error of the dual-branch design is much smaller than that of the  single-branch design.  
\vspace{-0.2cm}
\section{Conclusion and Discussion} \label{summary} \vspace{-0.1cm} 
We proposed IDEA-Net, 
an end-to-end framework for temporally interpolating dynamic 3D point cloud sequences with large non-rigid deformation. 
We formulated the problem as estimation of point-wise trajectories which can be approximated in a coarse-to-fine manner with the aid of the explicitly learned temporal consistency. By extensive experiments and ablation studies, we demonstrated the significant advantages of the proposed IDEA-Net over state-of-the-art methods in both quantitatively and visually. 
We believe our framework can provide novel insights for the acquisition and processing of dynamic point cloud sequences.

Despite this paper reveals essential discoveries on the problem of dynamic 3D point cloud interpolation, 
we can expect to further improve performance by enriching specific technical roadmap from different aspects. First, we can increase the receptive field of the time domain by feeding a longer frame group instead of a pair. Second, we can consider introducing sequential modeling framework to jointly learn temporal correlation across multiple frames. Third, considering the fact that points move at different speed and acceleration, we can design higher-order trajectory estimation schemes to empower the whole model. Finally, it is highly desired to construct a quantitative metric for reliably evaluating the quality of interpolated point cloud sequences.



\pagebreak

\onecolumn

\begin{center}
  \textbf{\Large IDEA-Net: Dynamic 3D Point Cloud Interpolation \\[.05cm]
  via Deep Embedding Alignment\\[.15cm]
  (Supplementary Materials)}\\[.4cm]
  
    Yiming Zeng$^{1}$\quad Yue Qian$^1$\quad Qijian Zhang$^1$\quad Junhui Hou$^{1}$\quad Yixuan YUAN$^1$ \quad Ying He$^2$\\
    $^1$
    City University of Hong Kong~~ 
    $^2$
    Nanyang Technological University\\
    {\tt 
    ym.zeng
    @my.cityu.edu.hk, 
    jh.hou
    @cityu.edu.hk  
    }\\[1cm]
\end{center}

\setcounter{section}{0}
\setcounter{equation}{0}
\setcounter{figure}{0}
\setcounter{table}{0}
\setcounter{page}{1}
\renewcommand{\thesection}{S\arabic{section}}
\renewcommand{\theequation}{S\arabic{equation}}
\renewcommand{\thefigure}{S\arabic{figure}}
\renewcommand{\thetable}{S\arabic{table}}
In this supplementary material, we first provide a demo video to 
comprehensively show the interpolation results in Section~\ref{sec-demo}. 
Then, we elaborate on the implementation details and training strategy of our IDEA-Net in Section~\ref{sec-impl}. 
We provide the details of the dataset used in our experiments in Section~\ref{sec-data}.
In Section~\ref{sec-abl}, we 
visualize the learned point-wise temporal consistency 
(i.e., matrix $\mathbf{A}$) by the dual- and single- branch models.
In Section~\ref{more-vis}, we provide more visual comparisons.

\section{Video Demo}\label{sec-demo}
We refer the readers to the video demo at \href{https://github.com/ZENGYIMING-EAMON/IDEA-Net.git}{https://github.com/ZENGYIMING-EAMON/IDEA-Net.git}, where we show the interpolation results ($k_{\text{train}}=k_{\text{test}}=3$) 
of our method and the compared methods.

\section{Implementation Details of IDEA-Net}\label{sec-impl}
We provide the implementation details in Table~\ref{tab:implementation}. We omit the BatchNorm layers and ReLU activation functions that are carried in each Conv layer, except for the Conv layer at the end of each module. In the feature representation module, we replace ReLU with LeakyReLU (negative slop is 0.2).

\begin{table}[htbp]
\renewcommand\arraystretch{1.50}
\centering
\caption{Implementation details of our network structure. The input and output dimensions are in parentheses.}
\label{CorrNet3D_detail}
\footnotesize
\setlength{\tabcolsep}{0.20cm}{
\begin{tabular}{c | c | c}
\toprule[1.0pt]
\textbf{Module} & \textbf{Layer / Operation} & \textbf{Output} \\ 
	\hline\hline
	& getGraphFeat (3, 6), Conv2d (6, 64), MaxPooling & x1 \\
	& getGraphFeat (64, 128), Conv2d (128, 64), MaxPooling & x2 \\
	& getGraphFeat (64, 128), Conv2d (128, 128), MaxPooling & x3 \\
	& getGraphFeat (128, 256), Conv2d (256, 256), MaxPooling & x4 \\
	Feature Embedding & Concat (x1+x2+x3+x4, 512), Conv1d (512, 512) & x5 \\
	& MaxPooling & x6 \\
	& AvgPooling & x7 \\
	& Concat (x6+x7, 1024) & y1 \\
	& Concat(x5+y1, 1536), MLP(1536, 512, 256, 128) & y2 \\
	\hline
	& $\mathbf{P_0},\mathbf{P_1}$$\rightarrow$shared feature embedding & $\mathbf{F_0}$, $\mathbf{F_1}$ \\
	Temporal Consistency & $\mathbf{F_0}$, $\mathbf{F_1}$ $\rightarrow$inverse distance & $\widetilde{\mathbf{A}}$ \\ 
	& $\widetilde{\mathbf{A}}$ $\rightarrow$ Conv1d (1024, 1024 + 128) & -\\
	& $\rightarrow$ row normalization & $\mathbf{A}$\\
	\hline
	& $\mathbf{P_0},\mathbf{P_1},\mathbf{F_0},\mathbf{F_1}$ $\rightarrow$ aligned by $\mathbf{A}$  & - \\
	& $\rightarrow$ linear interpolation & $\mathbf{P}_{0\rightarrow t}, \mathbf{F}_{0\rightarrow t},\mathbf{P}_{1\rightarrow t}, \mathbf{F}_{1\rightarrow t}$ \\
	Trajectory Compensation & $\mathbf{F}_{0\rightarrow t},\mathbf{F}_{1\rightarrow t}$ $\rightarrow$ $\{$Conv1d(1024+128, 1152, 576, 288, 3), Tanh$\}$ & $\mathbf{\Delta}_{0\rightarrow t}, \mathbf{\Delta}_{1\rightarrow t}$ \\
	& $\mathbf{\Delta}_{0\rightarrow t}, \mathbf{\Delta}_{1\rightarrow t}$ $\rightarrow$ compensate $\mathbf{P}_{0\rightarrow t},\mathbf{P}_{1\rightarrow t}$  & $\mathbf{O}_{0\rightarrow t}, \mathbf{O}_{1\rightarrow t}$\\    
\bottomrule[1.5pt]    
\end{tabular}}
\label{tab:implementation}
\end{table}

During the training process, we use the EMD loss for both branches and take the average of them as the final training loss. 
We use Pytorch~\cite{pytorch_2019} to implement our model on a single GPU (GeForce RTX 3090). 
We use Adam with the initial learning rate 0.0001 to optimize the model. 
For all the models, we train 1000 epochs with the batch size set to 14. 
For the feature embedding module in the table, we follow the original settings of DGCNN~\cite{wang2019dynamic}.

\section{Details of Constructed Datasets}\label{sec-data}
In this section, we provide more details for the DHB dataset and DFAUST dataset used in our paper. 
As shown in Table~\ref{tab:dataset}, both datasets are composed of 14 sequences, 
in which the four sequences (\textit{Longdress, Loot, Redandblack, Soldier}) are 3D point clouds, 
and others are 3D meshes. We uniformly sampled 1024 points from each individual frame to construct dynamic point cloud sequences. For the DHB dataset, 
we used the top 8 sequences to form the training set and the remaining as the testing set. For the DFAUST dataset, 
we used 
used the top 11 sequences 
for training and the rest for testing. All the datasets listed in the table will be released along with our code.

\begin{table}[htbp]
\centering
\caption{\label{Tab_DHB} Details of the DHB dataset and DFAUST dataset in our experiment.} 
\begin{tabular}{c|c||c|c}
\toprule 
\text { DHB sequences }  & \text {\# frames} & \text { DFAUST sequences }  & \text {\# frames}\\
\hline \text { Bouncing }  & \text { 175 } & \text { chicken\_wings }  & \text { 216 }\\
\hline \text { Crane }   & \text { 175 } & \text { hips } & \text { 697 }\\
\hline \text { Handstand }   & \text { 175 } & \text { jiggle\_on\_toes }  & \text { 240 }\\
\hline \text { Jumping }   & \text { 150 } & \text { jumping\_jacks }  & \text { 461 }\\
\hline \text { March\_1 }   & \text { 250 } & \text { knees } & \text { 500 }\\
\hline \text { March\_2 }   & \text { 250 } & \text { light\_hopping\_loose } & \text { 234 }\\
\hline \text { Samba }   & \text { 175 } & \text { light\_hopping\_stiff } & \text { 214 }\\
\hline \text { Squat\_1 }   & \text { 250 } & \text { one\_leg\_jump }& \text { 541 } \\
\hline \text { Squat\_2 }   & \text { 250 } & \text { one\_leg\_loose }& \text { 264 } \\
\hline \text { Swing }   & \text { 150 } & \text { punching }& \text { 303 } \\
\hline \text { Longdress }   & \text { 300 }& \text { running\_on\_spot } & \text { 331 } \\
\hline \text { Loot }   & \text { 300 } & \text { shake\_arms }& \text { 230 } \\
\hline \text { Redandblack }   & \text { 300 } & \text { shake\_hips }& \text { 243 } \\
\hline \text { Soldier }  & \text { 300 } & \text { shake\_shoulders }& \text { 255 } \\
\bottomrule
\end{tabular}
\label{tab:dataset}
\end{table}


\section{
Visual Illustration of the Learned Point-wise Temporal Consistency }\label{sec-abl}
\begin{figure}[htbp]
  \centering
  \subcaptionbox{\label{subfig_our} }{
  \includegraphics[width=0.75\linewidth]{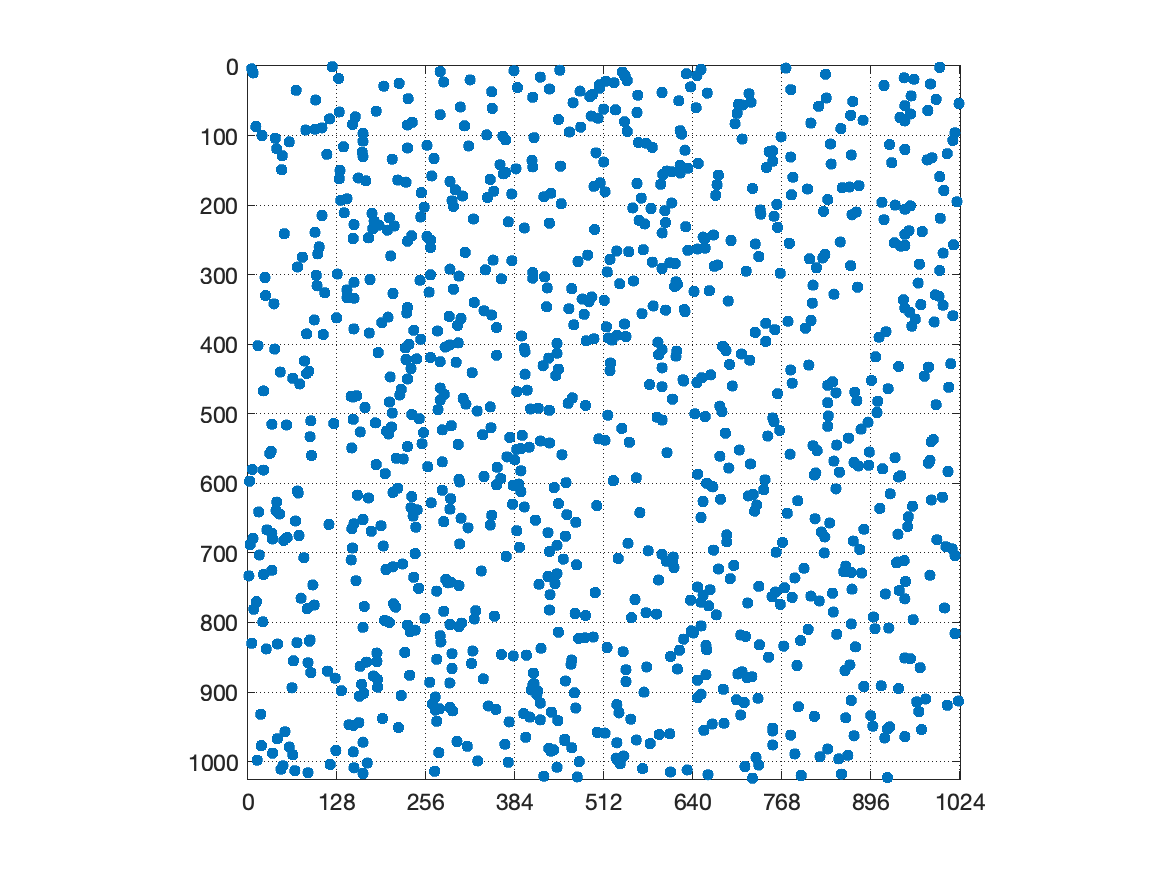}}
  
  \subcaptionbox{\label{subfig_our_branch} }{
  \includegraphics[width=0.75\linewidth]{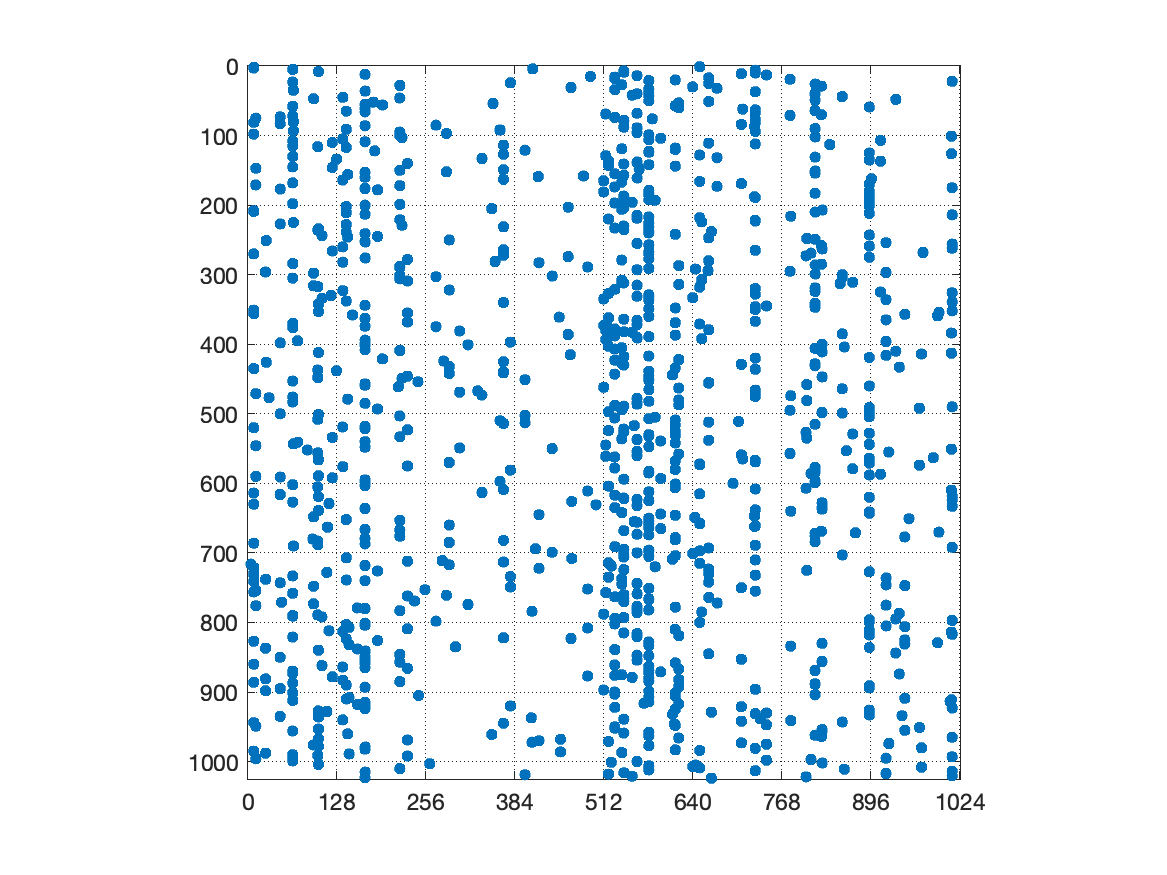}}
  
  \caption{Visual illustration of learned $\mathbf{A}$. (a) the dual-branch design. (b) the single-branch design.}
  \label{fig:matrix}
\end{figure}

As mentioned in Section \textcolor{red}{4.4(f)} of the manuscript, we conducted the ablation study to demonstrate the advantage of the dual-branch design over the single-branch design. Here, we also visualize the learned point-wise temporal consistency, (i.e., matrix $\mathbf{A}$) to illustrate the difference between the two designs. 
Keeping consistent with other ablation studies, we used the mixed data training mechanism, as mentioned in the last paragraph of Section \textcolor{red}{4.2} of the manuscript, to train and test our model on the DHB dataset and we fixed $k_{\text{test}}=3$. We fed a random sample pair of the sequence \textit{swing} into the network and obtained matrix $\mathbf{A}$. As shown in Fig.~\ref{fig:matrix},
it can be seen that the learned matrix $\mathbf{A}$ by the single-branch model has many values distributed in the same column (Fig.\ref{subfig_our_branch}), meaning that multiple points of $\mathbf{P}_0$ are aligned to an identical point of $\mathbf{P}_1$. However, such an observation rarely appear in the matrix $\mathbf{A}$ (Fig.\ref{subfig_our}) by the dual-branch model, which is credited to the regularization effect of the dual-branch design 
Note that the ground-truth point-wise consistency is not available, and thus we cannot quantitatively measure the accuracy of $\mathbf{A}$.   

\section{More Visual Results}\label{more-vis}
In this section, we illustrate more visual comparisons with the state-of-the-art 
flow-based method PointINet~\cite{lu2020pointinet} on the DHB dataset. 
As shown in Figs.~\ref{fig_OurPINETswing}, \ref{fig_OurPINETsoldier}, and \ref{fig_OurPINETsquat2}, we provide results of two methods on the sequences named \textit{swing}, \textit{soldier} and \textit{squat\_2}, respectively. Both methods were evaluated under the setting $k_{\text{train}}=3$ and $k_{\text{test}}=3$. It can be seen that for the sequence with large motion and rotations, i.e., \textit{swing} (Fig.~\ref{fig_OurPINETswing}), PointINet~\cite{lu2020pointinet} produces  broken limbs, which do not appear in our method. 
For the sequences with smooth motion, i.e., \textit{soldier} (Fig.~\ref{fig_OurPINETsoldier}) and \textit{squat\_2} (Fig. \ref{fig_OurPINETsquat2}), our method produces fewer artifacts and holes than PointINet~\cite{lu2020pointinet}.

\begin{figure}[htbp]
\centering
\subcaptionbox{}{
\includegraphics[width=0.49\linewidth]{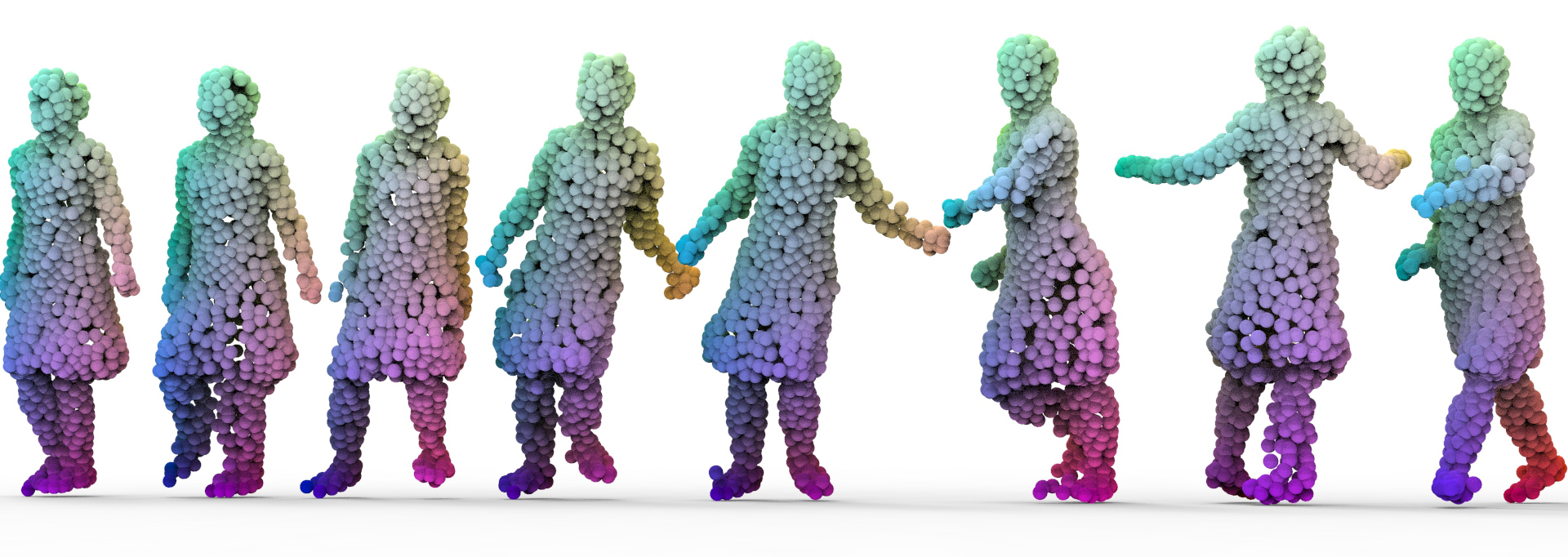}}
\label{OurPINETswing_a} 
\subcaptionbox{}{
\includegraphics[width=0.49\linewidth]{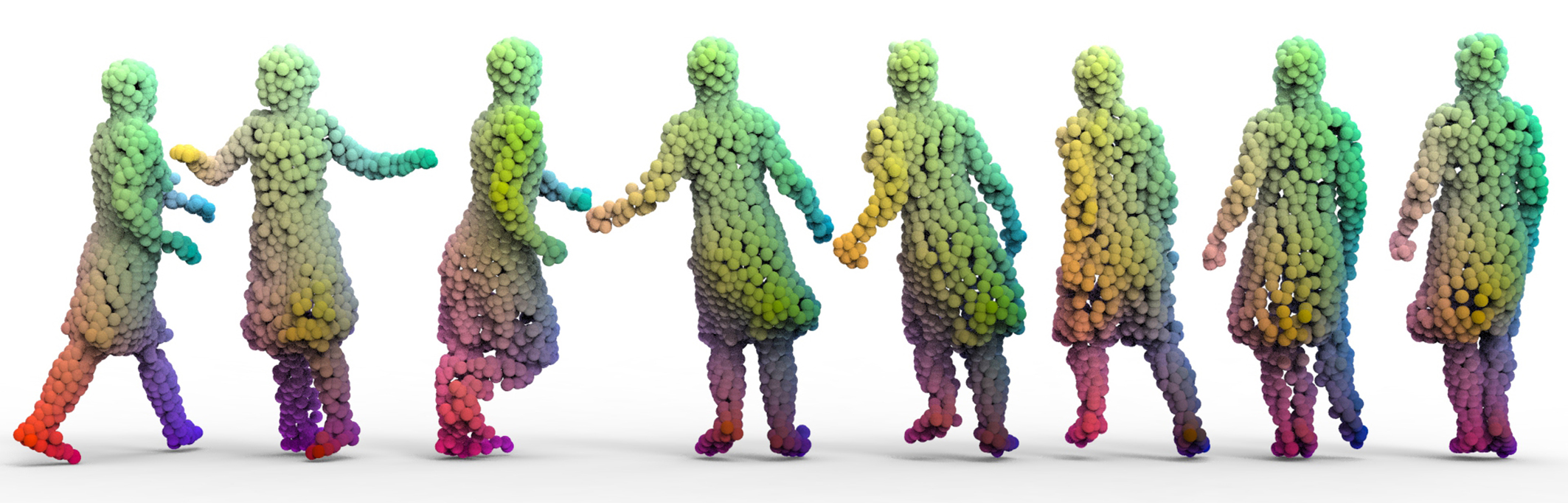}}
\label{OurPINETswing_b} 

\subcaptionbox{}{
\includegraphics[width=0.49\linewidth]{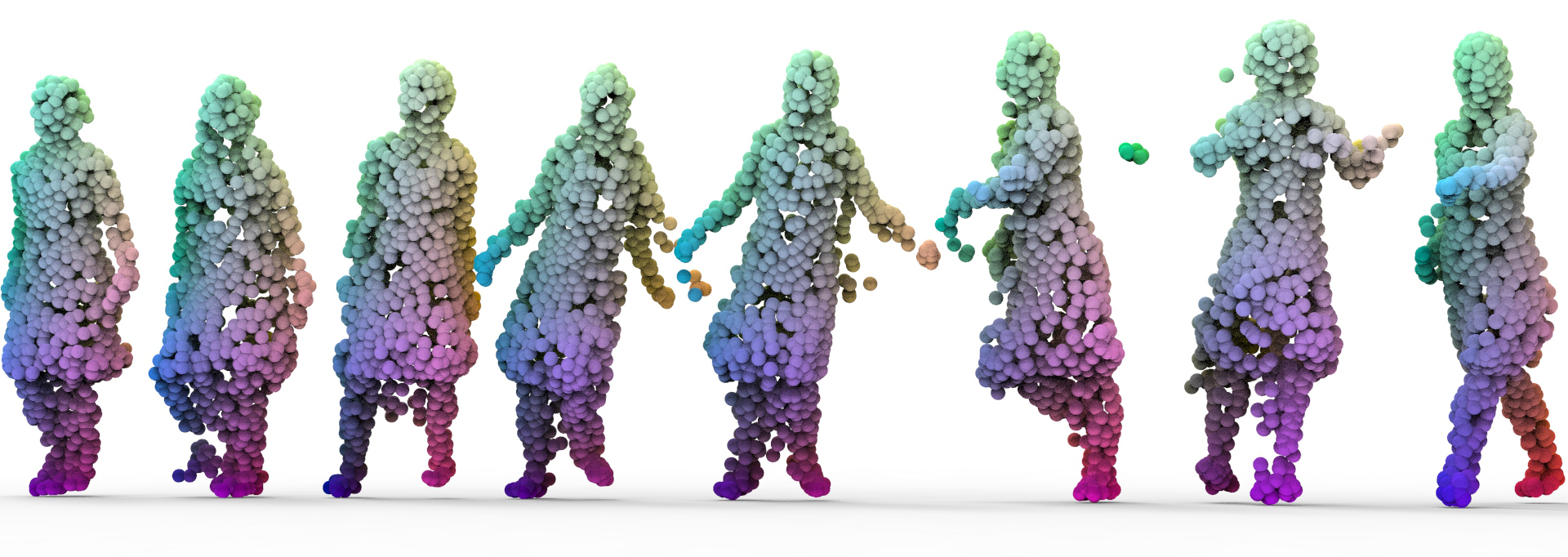}}
\label{OurPINETswing_c} 
\subcaptionbox{}{
\includegraphics[width=0.49\linewidth]{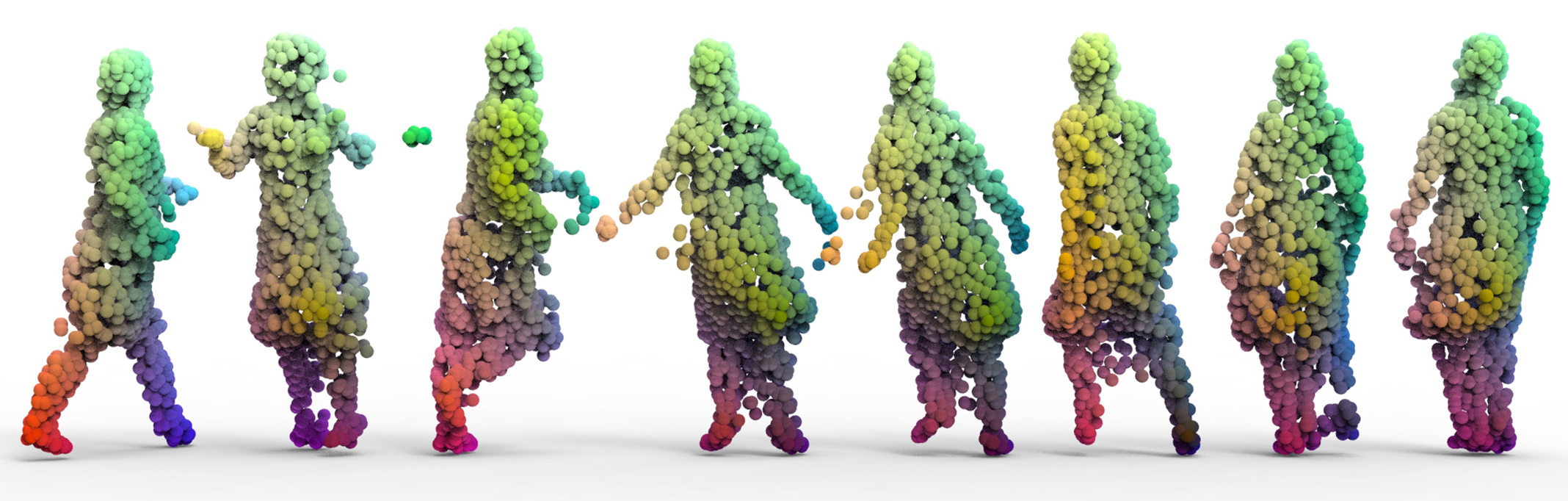}}
\label{OurPINETswing_d} 

\subcaptionbox{}{
\includegraphics[width=0.49\linewidth]{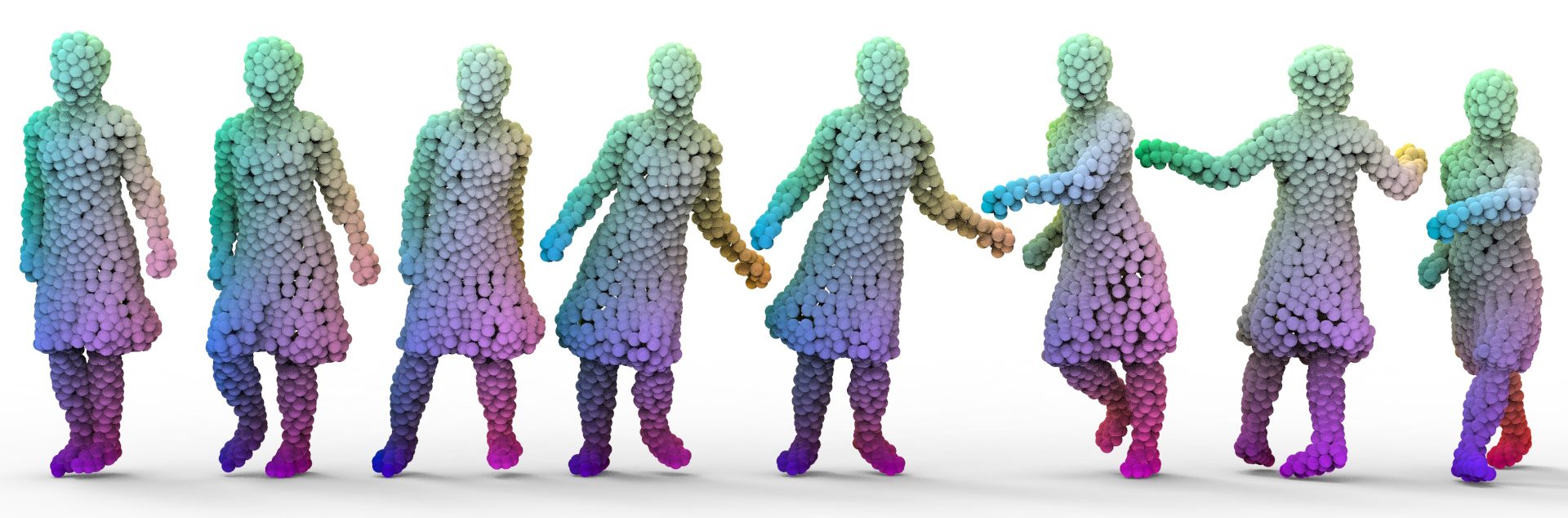}}
\label{OurPINETswing_e} 
\subcaptionbox{}{
\includegraphics[width=0.49\linewidth]{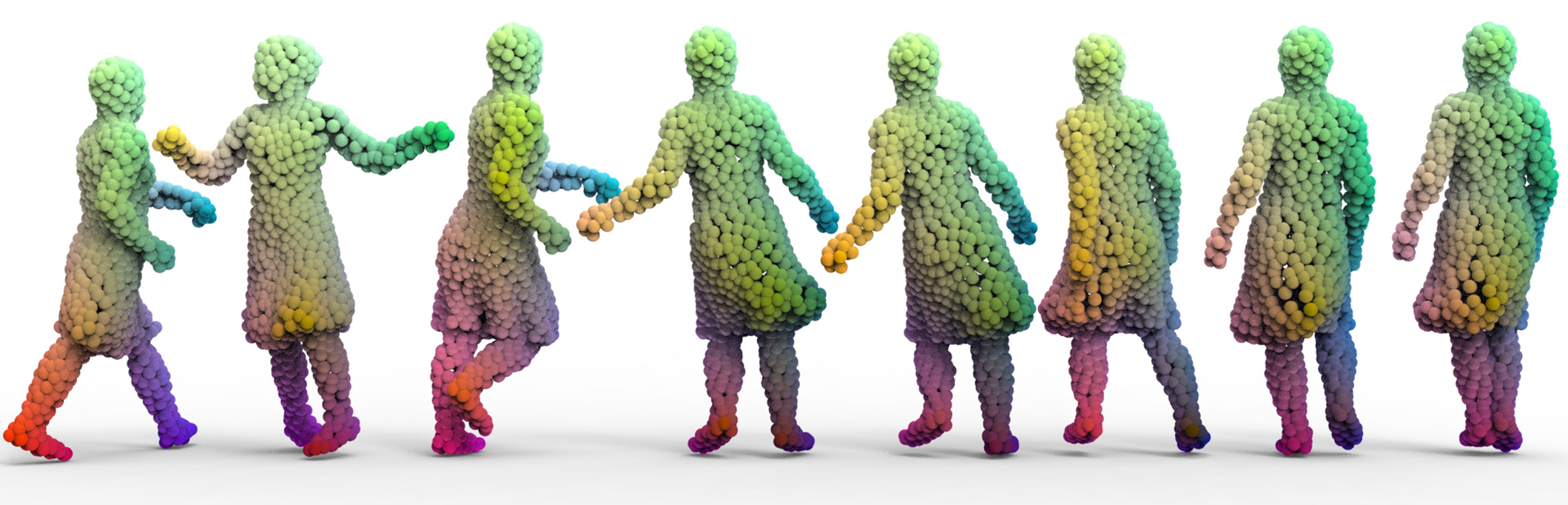}}
\label{OurPINETswing_f} 
\caption{Visual comparisons of the interpolated frames 
  on the test sequence \textit{Swing} by our method (the top row), PointINet (the middle row), and the ground-truth (the bottom row). (a), (c) and (e): front views; (b), (d) and (f): back views.} 
\label{fig_OurPINETswing} 
\end{figure}

\begin{figure}[htbp]
  \centering
  \includegraphics[width=0.9\linewidth]{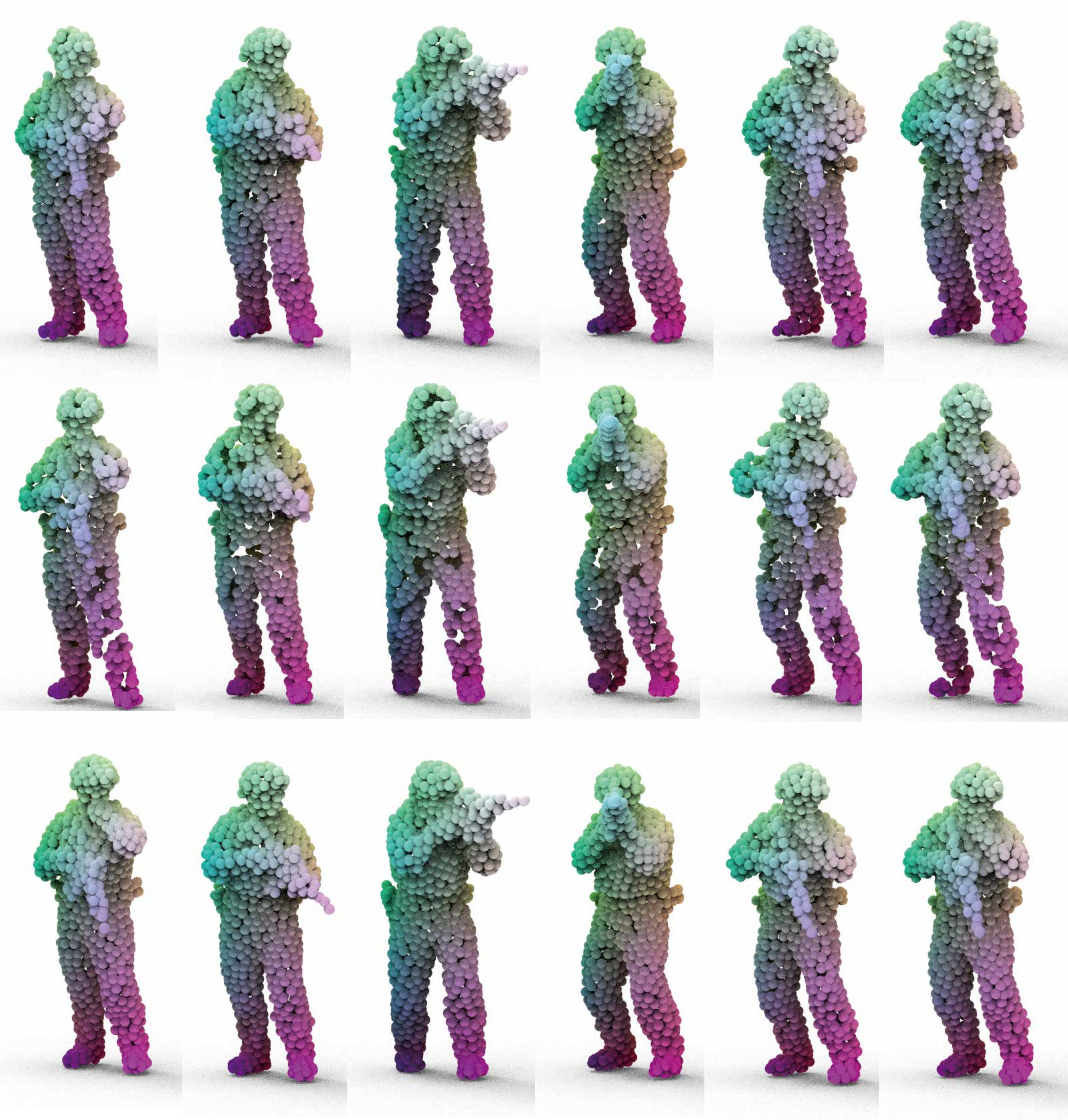}
  \caption{Visual comparisons of the interpolated frames 
  on the test sequence \textit{soldier} by our method (the top row), 
  PointINet (the middle row), and the ground truth (the bottom row). }
  \label{fig_OurPINETsoldier} 
\end{figure}

\begin{figure}[htbp]
  \centering
  \includegraphics[width=0.9\linewidth]{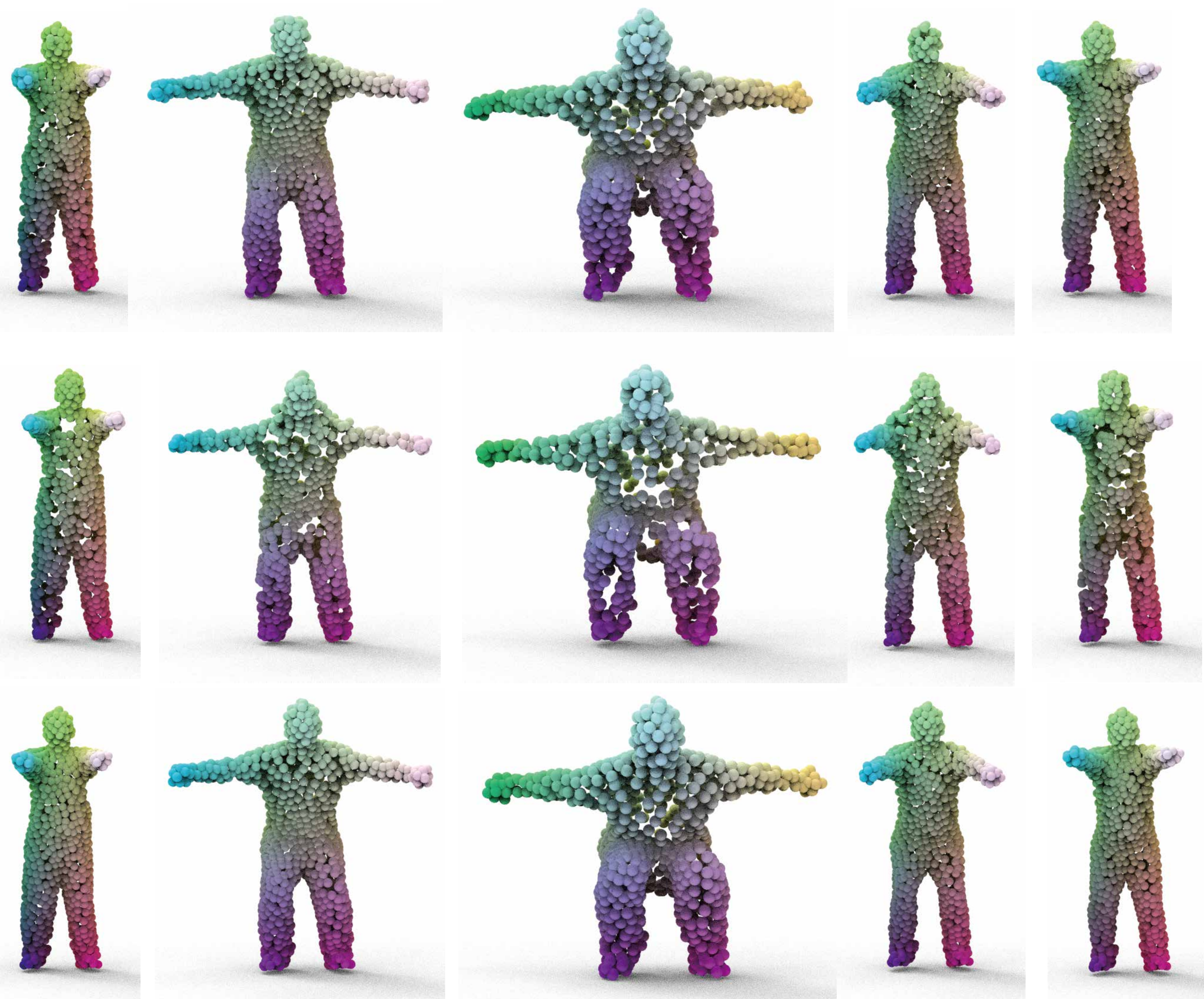}
  \caption{Visual comparisons of the interpolated frames 
  on the test sequence \textit{squat} by our method (the top row), 
  PointINet (the middle row), and the ground truth (the bottom row). }
  \label{fig_OurPINETsquat2} 
\end{figure}
\clearpage
{\small
\bibliographystyle{ieee_fullname}
\bibliography{egbib}
}

\end{document}